\pdfoutput=1
\documentclass{article} %
\usepackage[preprint]{iclr2027_conference}
\usepackage{times}

\usepackage{amsmath,amsfonts,bm}

\def\figref#1{figure~\ref{#1}}
\def\Figref#1{Figure~\ref{#1}}

\def\secref#1{section~\ref{#1}}

\def\eqref#1{equation~\ref{#1}}

\def\tabref#1{table~\ref{#1}}

\def\appref#1{appendix~\ref{#1}}

\def\1{\bm{1}}

\DeclareMathAlphabet{\mathsfit}{\encodingdefault}{\sfdefault}{m}{sl}
\SetMathAlphabet{\mathsfit}{bold}{\encodingdefault}{\sfdefault}{bx}{n}

\newcommand{\E}{\mathbb{E}}

\usepackage{hyperref}
\hypersetup{colorlinks=true, allcolors=blue,
  pdftitle={Make Every Expansion Count: Learning to Search Efficiently in LLM Code Repair},
  pdfauthor={Elad Sharony, Ido Greenberg, Jialin Song, Shie Mannor, Gal Chechik, Eli Meirom}}
\usepackage{url}
\usepackage{graphicx}
\usepackage{booktabs}
\usepackage{amsmath}
\usepackage{amssymb}
\usepackage{xcolor}
\definecolor{blue}{RGB}{28,106,137}
\usepackage{xspace}
\usepackage{flafter}
\usepackage{float}
\usepackage{tcolorbox}
\tcbuselibrary{listings}

\AddToHook{para/begin}{\looseness=-2\relax}

\definecolor{nvidia_green}{RGB}{118,185,0}

\newcommand{\ausc}{\operatorname{AUSC}}

\newcommand{\method}{CodeTriage}
\newcommand{\fmethod}{\method\xspace}

\newif\ifshowse    \showsetrue     %
\newif\ifshowtimes \showtimestrue  %
\newif\ifroundcost \roundcosttrue   %

\newcommand{\se}[1]{\ifshowse\textsuperscript{\textcolor{black!55}{$\pm#1$}}\fi}
\newcommand{\ci}[2]{\textsuperscript{\textcolor{black!55}{$[#1,\,#2]$}}}

\newcommand{\costhundredths}[1]{%
  \fpeval{floor(#1/100)}.%
  \ifnum\fpeval{#1-100*floor(#1/100)}<10 0\fi\fpeval{#1-100*floor(#1/100)}}

\newsavebox{\tabbox}

\newcommand{\modelicon}[1]{\raisebox{-0.15em}{\includegraphics[height=1.15em]{figures/icons/#1.pdf}}\hspace{0.35em}}

\title{Make Every Expansion Count: Learning to Search Efficiently in LLM Code Repair}

\author{%
  Elad Sharony\thanks{Corresponding author: \href{mailto:eladsharony@campus.technion.ac.il}{\nolinkurl{eladsharony@campus.technion.ac.il}}.} \\
  Technion, NVIDIA \\
  \And
  Ido Greenberg \\
  NVIDIA \\
  \And
  Jialin Song \\
  NVIDIA \\
  \AND
  Shie Mannor \\
  Technion, NVIDIA \\
  \And
  Gal Chechik \\
  NVIDIA \\
  \And
  Eli Meirom \\
  NVIDIA \\
}

\begin{document}

\maketitle

\begin{abstract}
Large language model (LLM) code-repair systems fix faulty programs by repeatedly generating and testing revisions.
Each revision is costly, making the choice of which repair to pursue important for efficiency. Existing search policies rely on test pass rates and search statistics, ignoring the details of test failures and proposed edits.
We introduce \textbf{\fmethod{}}, a learned policy that prioritizes edits before their revised programs are generated and tested. Its 0.6B-parameter model reads the test output, failure diagnosis, and proposed edit, then predicts a distribution over the number of repair steps needed to reach a fix through that edit. 
These predictions prioritize all pending edits across programs before generating and testing the revised code.
On held-out repair tasks, \fmethod{} finds fixes with \textbf{21\%} fewer program revisions and \textbf{12\%} fewer output tokens than the strongest baseline on each metric.
Without retraining, it achieves the highest average success across search budgets for each of 11 unseen repair LLMs.
Ablations show that both the failure context and edit text, as well as comparisons across programs, improve search.
We also release a corpus of 12.2 million program versions from four LLMs, organized into repair trees with proposed edits and test results.

\end{abstract}

\ificlrfinal
\vspace{-4pt}
{\centering Project page: \href{https://esharony.me/projects/codetriage/}{esharony.me/projects/codetriage}\par}
\vspace{-4pt}
\fi

\section{Introduction}
\label{sec:intro}

Coding agents have revolutionized how we write software and automate processes, but producing correct code remains a central challenge. A coding agent rarely fixes a difficult bug in one attempt. LLM-based repair systems often operate in a feedback loop: generate a program revision, run the tests, inspect the failures, and propose edits describing what to change and why \citep{chen2024teaching,shinn2023reflexion,olausson2024self}. This loop makes progress measurable, but it also makes repair expensive. Every attempted edit requires another program generation and test execution. Under a fixed test-time budget, success therefore depends not only on generating good repairs, but also on deciding which proposed repair should receive the next expensive attempt.

Committing to a single repair approach and repeatedly refining the same program can exhaust a limited budget without finding a fix. Exploring more alternatives could improve solution coverage \citep{brown2024large,snell2025scaling}, but applying each edit requires generating and testing a revised program. Prior approaches choose which program to refine next using the fraction of unit tests it passes and how often it has already been refined \citep{tang2024code,inoue2026wider,miyamoto2026aligning}.

Yet the fraction of passing tests does not reveal why a program fails or how difficult it will be to repair. Two programs with the same pass rate may be very different distances from a solution: one may have a single localized error, while the other passes the same number of tests despite several interacting faults. Search must also decide between repairs to different programs: one program's most promising edit may still be less promising than another's second-best. A search policy should therefore compare proposed edits across programs before the costly code generation and testing phase.

We introduce \fmethod{}, a learned search policy that uses a small 0.6B-parameter priority model to rank edit proposals before the revised programs are generated and tested. Given the test output, the failure diagnosis, and the text of a proposed edit, the model estimates the edit's likelihood of leading to a fix and the number of repair steps required. The repair LLM proposes and implements edits; the priority model only selects among them. This separation allows us to use the same priority model across different repair LLMs.

On complete repair searches (repair-trees), \fmethod{} finds fixes with \textbf{21\%} fewer program revisions and \textbf{12\%} fewer output tokens than the strongest baseline on each metric (\Figref{fig:main}). Without retraining, the same priority model transfers to eleven unseen repair LLMs and achieves the highest success rate averaged across budgets (\Figref{fig:transfer}). Matched ablations support the proposed mechanism: both parent-state text and proposed edits improve search, and learning to compare edits across the full frontier further reduces search cost (\tabref{tab:ablations}).

\noindent\textbf{Contributions.}
Our main contribution is a learned search policy that prioritizes proposed repairs across programs and repair depths before generating and testing revised code. It uses test output, failure diagnoses, and proposed edits to predict a distribution over the steps needed to reach a fix, providing a common basis for deciding which repair to pursue next. We further show that a small priority model trained on one LLM's repairs can guide other repair LLMs without retraining. Finally, we release a 12.2-million-node repair-tree corpus recording alternative repair paths and their outcomes to support further research on code-repair search.

\section{Problem Setting}
\label{sec:setting}

Our goal is to fix a faulty program while minimizing latency and token usage, which boils down to minimizing the number of generated programs and tests. A repair LLM, called the \emph{proposer}, reads the test output (\emph{execution feedback}), explains the failure, and describes candidate repairs as \emph{proposed edits}. Each proposed edit specifies what to change and why, before the revised program is generated.  

\begin{figure}[ht]
    \centering
    \includegraphics[width=\textwidth,
        alt={Diagram of a repair tree. The root program passes 30 percent of tests; its two expanded children pass 10 and 60 percent. Expanding the 60-percent child's proposed edit yields a program that passes every test. A side panel shows what an expanded node holds: the program, its pass rate, test feedback, a diagnosis, and k proposed edits. The unexpanded proposed edits across the tree form the shaded frontier.}]{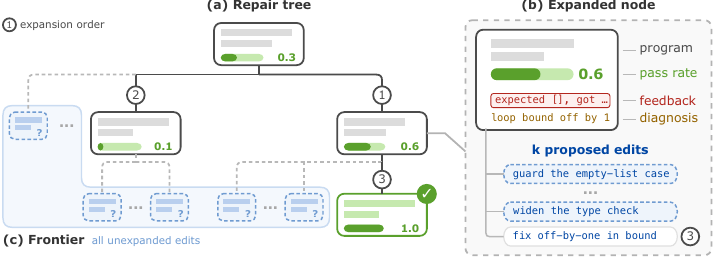}
    \caption{\textbf{The search problem.} \textbf{(a)} \textbf{Repair tree}: each expanded node proposes $k$ edits; the policy expands one edge at a time, in the circled order, until a node passes every test. \textbf{(b)} \textbf{Expanded node}: what the policy observes, i.e.\ the program, its pass rate, the execution feedback, the diagnosis, and the $k$ proposed edits. \textbf{(c)} \textbf{Frontier} (shaded): all proposed edits not yet expanded (dashed, blue); their programs have not been generated, so their pass rates are unknown.}
    \label{fig:setting}
\end{figure}

\paragraph{Code repair trees.}
\label{sec:trees}
Each repair tree begins with an initial program as its \emph{root} (\Figref{fig:setting}a).
Every \textbf{node} is a program executed against the task's full test suite, with its pass rate (fraction of tests passed) measured. A node solves the task if its pass rate is 1.0.
Each node stores its \textbf{execution feedback} (the output of running the tests) and a failure \textbf{diagnosis}. The proposer also supplies $k$ \textbf{proposed edits}, each describing what to change and why (\Figref{fig:setting}b). At this stage, each edge contains only the \textbf{proposal text}; the complete revised program has not yet been generated. Applying a proposed edit is costly, as it generates a new program. Hence, our cost unit is an edge \emph{expansion}: a program generation phase that also runs the full test suite and adds the evaluated result as a child node.
With maximum depth $D$ and $k$ proposed edits at each node, an exhaustive tree contains at most $\sum_{d=1}^{D} k^d$ non-root nodes, each corresponding to one possible expansion.

\paragraph{The search problem.}
\label{sec:contract}
The \emph{frontier} contains all proposed edits that have yet to be tried, including alternatives for programs explored earlier, i.e., every unexpanded outgoing edge of an expanded node (\Figref{fig:setting}c).  A search policy assigns a priority to each candidate (edge) in the frontier, guiding which repairs should be applied next. 

Expanding an edge creates and evaluates its child node. The child's program, pass rate, execution feedback, and diagnosis then become available, and its $k$ outgoing proposed edits are added to the frontier.
The objective is to minimize the number of expansions required to find the first solution. 

\paragraph{Proposal and search.}

The \emph{proposer} is an LLM that creates $k$ proposed edits from a node. The \emph{search policy} chooses which proposed edit to expand.  A search policy can only use the programs and feedback already observed, the proposed-edits text, and the search history, but not the unknown outcomes of unexpanded edges.
\fmethod{} is a \emph{search policy}. It doesn't change the \emph{proposer}.

We evaluate search policies on \emph{exhaustive trees}: precomputed repair trees in which every edge through depth $D$ has already been expanded.
This \emph{exhaustive-tree evaluation} replays each policy's search decisions on the same set of generated candidates and isolates the effect of expansion order.
We complement this evaluation with online search using eleven previously unseen proposers (\secref{sec:online}).

\section{\fmethod{}: Reading the Frontier as Survival Distributions}
\label{sec:method}

Exploring every proposed edit is costly because each expansion requires generating a revised program and running its unit tests. Under a limited budget, the policy must prioritize edits across programs and repair depths before their outcomes are known. \fmethod{} makes these comparisons by predicting a distribution over each edit's \textbf{solve distance}: the fewest successive expansions needed to reach a fix.

The model reads the proposed edit together with the \emph{parent state} to assess how well the proposed change addresses the observed failure. The parent state contains execution feedback, the failure diagnosis, and four numerical features: pass rate, normalized depth, failed-test fraction, and normalized program length. These inputs provide a compact representation of each repair candidate. The edge's remaining depth sets the horizon of the priority rule \eqref{eq:priority}.

\begin{figure}[t]
    \centering
    \includegraphics[width=\textwidth,
        alt={Pipeline diagram. Test feedback, a failure diagnosis, and one proposed edit are serialized and read by a 0.6B Qwen3 language model with LoRA adapters. Its output is concatenated with four numeric features and passed to an MLP that gives each edit a priority score. Best-first search expands the highest-scoring edit.}]{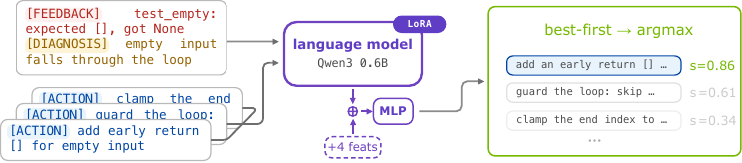}
    \caption{\textbf{\fmethod{}.} Left to right: the parent feedback, diagnosis, and proposed edit for each frontier edge are serialized into one sequence and passed through a 0.6B language model. Four numeric features are concatenated ($\oplus$) before the
    MLP head emits a $(D{+}1)$-bin distribution over the edge's \emph{solve distance} $t$,
    the number of expansions its branch needs to reach a fix. That distribution reduces to
    one scalar priority $s$, and deterministic best-first search
    expands the argmax (green).}
    \label{fig:method}
\end{figure}

\paragraph{Solve distance and the censored objective.}
\label{sec:objective}
For an edge $e=(v,u)$, its solve distance $t(e)$ counts the expansion of $e$ plus the shortest path from the child $u$ to an observed solution.
Thus $t(e)=1$ when $u$ solves the task; otherwise, $t(e)=1+\min_{e'\in\operatorname{out}(u)}t(e')$, where the minimum is over child edges whose subtrees contain a solution.
If $u$ is at depth $d$ in a tree of maximum depth $D$, then $b=D-d+1$ is the largest solve distance observable for $e$.

The model outputs probabilities $(p_1,\ldots,p_D,p_{>D})$, where $p_j=P(t=j)$ and $p_{>D}=P(t>D)$, and $F(m)=\sum_{j\leq m}p_j$ is the cumulative probability that the solve distance is at most $m$.
When a solution is observed within the remaining depth $b$, the exact target is $t=t(e)$.
Otherwise, the tree reveals only that $t>b$; this is a \emph{censored} observation.
Treating a censored branch as permanently unsolvable would discard the possibility of a later fix. Instead, the likelihood sums the probability mass beyond $b$, following standard discrete-survival likelihoods \citep{tutz2016modeling}.
Training therefore uses
\begin{equation}
    \mathcal{L} = \begin{cases}
        -\log p_t, & \text{observed distance } t,\\[2pt]
        -\log\bigl(1 - F(b)\bigr), & \text{censored at remaining depth } b.
    \end{cases}
    \label{eq:loss}
\end{equation}

\paragraph{From distribution to search priority.}
\label{sec:priority}
The priority should favor edits that are likely to lead to a fix within the remaining depth and require fewer repair steps. The probability of finding a fix alone does not distinguish short repair paths from long ones. We therefore average the predicted probabilities of finding a fix within each available number of steps. This gives the normalized area under $F$ up to the remaining depth $b$:
\begin{equation}
    s(e) = \frac{1}{b}\sum_{m=1}^{b} F(m),
    \label{eq:priority}
\end{equation}
Here $F(m)$ is the probability that the edit has a successful repair path of at most $m$ expansions.
Averaging these probabilities gives more weight to fixes predicted sooner and no weight to distances beyond the remaining depth.
Dividing by $b$ keeps scores on the same $[0,1]$ scale across depths.
Search is deterministic: at each step, it expands the frontier edge with the largest score.

\paragraph{Priority-model architecture.}
\label{sec:architecture}
The policy scores every proposed edit, so the cost of scoring must be negligible relative to program generation and unit-test execution. We therefore use a small 0.6B-parameter decoder-only language model that reads the parent's execution feedback, diagnosis, and proposed edit as a single sequence (\Figref{fig:method}). Its last-token hidden state is concatenated with four parent-level numeric features---pass rate, normalized depth, failed-test fraction, and normalized program length---and passed to an MLP head that predicts the $(D{+}1)$-bin solve-distance distribution. The architecture and training configuration are specified in \secref{sec:model}.

\section{Experimental details}
\label{sec:experiments}

\paragraph{Dataset.} We study approximately 8,700 Python programming problems from the \texttt{hard} subset of KodCode-V1 \citep{xu2025kodcode}. This spans problems from Codeforces, TACO, Code Contests, and APPS \citep{codeforces,li2023taco,li2022competition,hendrycks2021measuring}. We construct the main corpus with Qwen3-Coder-30B-A3B-Instruct \citep{yang2025qwen3}, hereafter Qwen3-Coder-30B, as the \emph{proposer}. To measure how the choice of proposer LLM affects search performance, we used three additional LLMs as proposers to build a corresponding auxiliary corpus:  Devstral-Small-2 \citep{mistral2025devstral2}, Qwen3-Coder-Next \citep{qwen2026qwen3codernext}, and Nemotron-Cascade2 \citep{yang2026cascade}. 

\label{sec:dataset}
Our dataset consists of exhaustive repair trees that record the proposed edits, the programs they produce, and the test results for those programs. We retain trees with at least one reachable fix. For each proposed edit with a successful repair path, the tree gives its solve distance: the fewest edits needed to reach a fix, including that edit. Branches cut off at the depth limit provide lower bounds. These trees are used to compare search policies. Together, the auxiliary and main corpora contain 12.2 million executed nodes. Details are provided in \appref{app:dataset}.

We split the data by programming problem rather than by repair tree. All trees derived from a problem belong to the same train, validation, or test split. An unseen problem in our results was not used for training or validation.

\paragraph{Metric.}
\label{sec:metric}
Let $T_i^\pi$ denote the expansion at which policy $\pi$ first finds a solution for root $i$.
Every tree in the exhaustive evaluation set contains a solution, so our primary cost, the mean number of expansions to a solution $\E[T^\pi]$ (lower is better), is finite.
Under a limited expansion budget $B$, unsuccessful runs do not reveal $T_i^\pi$. We therefore report success $S_\pi(B)=P(T^\pi\le B)$ and realized cost $C_\pi(B)=\E[\min(T^\pi,B)]$, where an unsuccessful run consumes the full budget. 
We summarize performance across the full budget range by the normalized area under the success curve,
\(\ausc_\pi(B)=\frac{1}{B}\sum_{j=1}^{B}S_\pi(j)\).

For the main test comparison, we train \fmethod{} with multiple seeds, select a single checkpoint based on validation performance, and then evaluate it on the sealed test set. Stochastic baselines are averaged over inference-RNG seeds. 
For each ablation, we report the average over three matched training seeds and include a full-model reference trained with the same seeds. Pairing seeds supports comparisons of each change while measuring variation across training runs. All test uncertainties are bootstrap standard errors clustered by the 982 source tasks.

\paragraph{Baselines.}
\label{sec:baselines}
We compare with \emph{random}, uniform over the frontier;
\emph{breadth-first search} (BFS), shallowest-first;
\emph{depth-first search} (DFS), deepest-first;
\emph{pass-rate}, always expand a random unexpanded edge of the highest-pass-rate expanded node (pure exploitation on observed outcomes).
We also compare with REx \citep{tang2024code}, both posterior variants of AB-MCTS-A \citep{inoue2026wider} (hereafter AB-MCTS), and BG-MCTS \citep{miyamoto2026aligning}. All policies search the same proposed edits, isolating how they allocate expansions. REx and BG-MCTS are tuned on validation, while AB-MCTS uses its published conjugate priors. We report the stronger AB-MCTS variant for each metric unless specified otherwise; \appref{app:baselines} describes the adaptations and tuning, and \appref{app:perseed} reports both variants.

\paragraph{Priority model and training.}
\label{sec:model}
The priority model is Qwen3-0.6B-Base \citep{yang2025qwen3}, trained with low-rank adaptation (LoRA) \citep{hu2021lora}; its base weights remain frozen.
We optimize the LoRA adapters and the prediction head with AdamW at constant learning rates of $8\times10^{-5}$ and $10^{-5}$, respectively, and select checkpoints based on validation search cost. Full training details are in \appref{app:training}.

\paragraph{Evaluation protocol.}
We evaluate policies in two settings. Offline replay uses 5,001 exhaustive Qwen3-Coder-30B trees from 982 held-out tasks, each containing a fix within the maximum repair depth. Every policy searches the same precomputed edits, with outcomes revealed only upon expansion. For transfer, we train and select the priority model using only Qwen3-Coder-30B data, then deploy it unchanged with eleven unseen repair LLMs. This evaluation retains starting programs regardless of solvability and weights each proposer equally.

Ablations and diagnostics use a separate validation split of 4,996 trees. Full protocol details and per-proposer results appear in \appref{app:online}.

\section{Results}

\subsection{Fewer Expansions and Output Tokens}
\label{sec:results}
\label{sec:main}

\begin{figure}[ht]
    \centering
    \includegraphics[width=\textwidth,
        alt={Two bar charts comparing eight search policies on held-out repair trees. CodeTriage needs the fewest mean expansions to the first fix, 11.54 against about 14.6 for the next best policy, AB-MCTS. It also needs the fewest mean output tokens, about 15.0 thousand against about 17.0 thousand for the next best policy, REx.}]{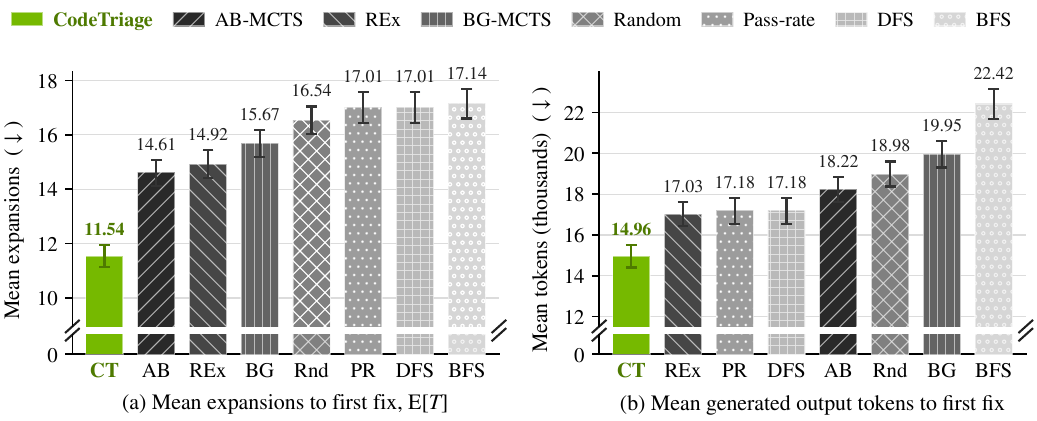}
    \caption{\textbf{Exhaustive-tree evaluation on unseen Qwen3-Coder-30B trees.}
    Mean \textbf{(a)} expansions and \textbf{(b)} output tokens through the first solution.
    Error bars show standard errors.
    Paired differences on the same roots are reported in \tabref{tab:id-full}.}
    \label{fig:main}
\end{figure}

We first measure the number of expansions and output tokens each policy uses to find a fix. Each policy searches the same held-out repair trees, so the program revisions and test outcomes are fixed. \Figref{fig:main} compares these costs.

\fmethod{} uses 21\% fewer expansions than AB-MCTS, the strongest baseline (11.54 compared with 14.59 on average). It uses 30\% fewer expansions than random search (11.54 compared with 16.54; \Figref{fig:main}a). Across the evaluated expansion budgets, \fmethod{} achieves higher success than every baseline at each matched mean expansion cost (see the first panel of \Figref{fig:pareto-all} in the appendix).

The reduction in expansions also translates into fewer generated output tokens. On the same 5,001 test roots, \fmethod{} uses 12\% fewer output tokens than REx, the most token-efficient baseline, and 21\% fewer than random search (\Figref{fig:main}b; \tabref{tab:tokens-full}). \appref{app:priority-cost} separately quantifies priority-model overhead, and \appref{app:perseed} gives the complete results.

\subsection{Transfer to Unseen Repair LLMs}
\label{sec:online}

\begin{figure}[ht]
    \centering
    \includegraphics[width=\textwidth,
        alt={Left: success rate against mean expansions for budgets from 4 to 36, averaged over eleven unseen repair LLMs; the CodeTriage curve lies above REx, AB-MCTS, BG-MCTS, DFS, and BFS. Right: for each LLM, the gain in area under the success curve over its strongest baseline, with 95 percent intervals. Gains range from 0.01 to 2.13 percentage points, with a mean of 0.90.}]{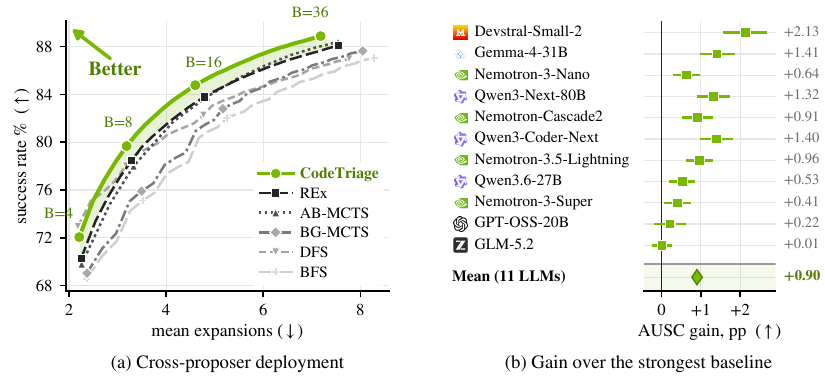}
    \caption{\textbf{Transfer to eleven unseen repair LLMs without retraining.}
    \textbf{(a)} Success rate versus mean expansions for budgets 4--36, averaged equally over proposers; failures use the full budget. (\appref{app:online}).
    \textbf{(b)} $\ausc$ gain over each proposer's strongest baseline, in percentage points, ordered by baseline $\ausc$. Bars show 95\% task-clustered paired-bootstrap intervals; the diamond marks the mean.}
    \label{fig:transfer}
\end{figure}

We evaluate the same priority model, trained on Qwen3-Coder-30B repair trees, with eleven unseen repair LLMs, without retraining or proposer-specific tuning. This experiment tests whether the learned selection policy transfers across coding LLMs. \Figref{fig:transfer}a reports performance across budgets, and \Figref{fig:transfer}b compares results for each proposer.

Averaged across the eleven unseen proposers, \fmethod{} matches or exceeds every baseline's success rate at every reported budget. \fmethod{} achieves these success rates at an equal or lower mean expansion cost (\Figref{fig:transfer}a). For example, at a budget of 16 expansions, \fmethod{} solves 84.8\% of problems compared with 83.8\% for REx, a gain of 1.0 percentage point. \fmethod{} uses 4\% fewer expansions on average: 4.60 compared with 4.79 for REx (\tabref{tab:online-mean}).

To summarize performance across budgets of 1--36 expansions, we report $\ausc(36)$, the normalized area under the success curve. For each coding LLM, we compare \fmethod{} with the best baseline (highest $\ausc(36)$). \fmethod{} has the higher score for all eleven proposers, with gains of up to 2.13 percentage points. The gains are statistically significant on nine proposers (\Figref{fig:transfer}b).  Averaged across proposers, \fmethod{} improves $\ausc(36)$ from 81.60 to 82.50, a gain of 0.90 percentage points. Per-proposer scores (\tabref{tab:online}), full budget sweeps, and the paired-bootstrap protocol are reported in \appref{app:online}.

\subsection{Ablations}
\label{sec:ablations}

We examine three design choices: which text the policy reads, how it learns to compare candidates, and how its predictions determine search priority. See  \appref{app:mechanism} and \appref{app:reductions} for details.

\paragraph{Parent state and proposed edits both matter.}
\label{sec:signal}
We ablate parent-state text (test output and diagnosis) and proposed-edit text use (\tabref{tab:ablations}).
Removing parent text increases cost by 16\% (from $11.80$ to $13.64$ expansions). Removing the proposed-edit text increases cost by 22\% (to $14.38$). Removing both increases cost by 45\% (to $17.14$). The model with both text inputs uses 31\% fewer expansions than the matched no-text policy, saving $5.35$ expansions.
The same conclusion holds on trees generated from Devstral-Small-2, Nemotron-Cascade2, and Qwen3-Coder-Next: relative to the matched no-text model, adding either text cuts cost by 14--21\% (1.3--2.1 expansions), and using both inputs is best for every proposer (\appref{app:ood-inputs}).

To assess how much the text encoder contributes, we compare \fmethod{} with a simpler policy trained on TF--IDF word-frequency features on the same dataset. \fmethod{} requires 16\% fewer expansions ($11.79$ versus $13.97$), showing that its gains extend beyond what these simple text features capture (\appref{app:cheap-baseline}).

\paragraph{Comparing edits across programs improves search.}
\label{sec:target}
\fmethod{} compares edits across programs and repair depths by predicting probabilities over the length of a shortest successful repair path. To examine the value of these comparisons, we train two models that assign a single score to each edit. A \emph{sibling ranker} learns to compare edits proposed for the same program. A \emph{global ranker} compares edits available together during search, including edits from different programs and depths. Both use the same pairwise ranking loss, with depth-aware targets for global comparisons (\appref{app:global-ranking}).

Global ranking reduces mean search cost from $13.32$ to $12.20$ expansions, recovering 73\% of the sibling ranker's gap to \fmethod{}. \fmethod{} achieves the lowest mean cost among the tested objectives, at $11.79$ expansions (\tabref{tab:objectives-full}). These results show the value of comparing edits across the full search frontier and the effectiveness of solve-distance predictions as a common basis for prioritizing repairs.

\paragraph{Rank by \emph{when} a fix arrives, not just \emph{whether}.}
\label{sec:reduction}
We hold the trained checkpoints fixed and change only the mapping from the predicted solve-distance distribution to a priority, isolating the effect of the search rule (\tabref{tab:ablations}; \appref{app:reductions}).
Using only the probability of finding a solution within the remaining depth, $F(b)$, raises cost by 4\%, from $11.79$ to $12.23$ expansions.
Using only the probability that the next expansion solves the task, $p_1$, raises cost by 15\%, to $13.58$. Both immediate and later solutions contribute useful information, and solutions predicted to occur sooner should receive more weight.

Normalization by remaining depth also matters. Removing the division by $b$ from \eqref{eq:priority} increases the cost by 8\% to $12.71$ expansions.
Replacing the scores with ranks computed separately at each depth raises cost by 18\%, to $13.96$ (\appref{app:reductions}). Both alternatives perform worse than the proposed score, which supports comparisons across depths.

\subsection{Search Performance Across Training-Data and Model Sizes}
\label{sec:scaling}
\label{sec:backbones}

\begin{figure}[ht]
    \centering
    \includegraphics[width=\textwidth,
        alt={Left: mean expansions to the first fix fall steadily, from about 14.7 to 11.9, as the training set grows from 1 to 100 percent, following a fitted power law. Right: mean expansions against backbone size for SmolLM2, Llama-3.2, Qwen3, Qwen3.5, and SmolLM3. Larger backbones search more efficiently, and Qwen3.5 is lowest among models of similar size.}]{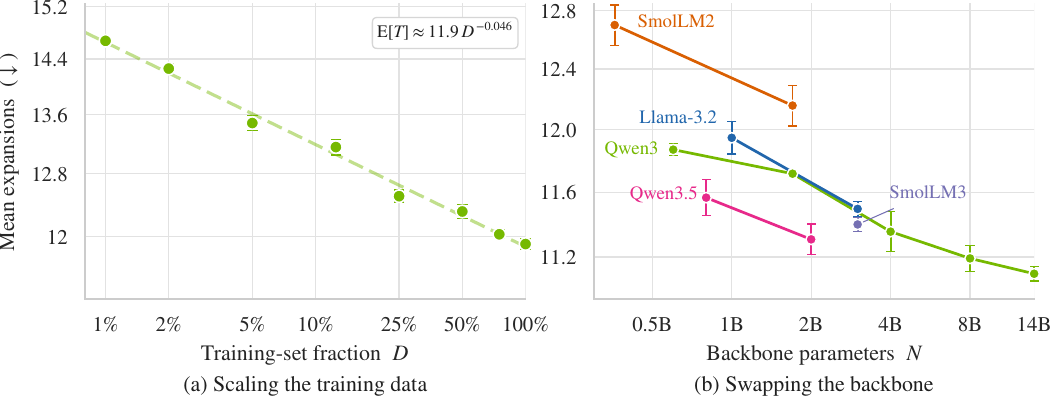}
    \caption{\textbf{Scaling and backbone sweeps on validation.} Lower $\E[T]$ is better. \textbf{(a)} Training-data scaling over a 100$\times$ range. \textbf{(b)} The same training recipe across five backbone families; Qwen3 provides the within-family size sweep. Points average three training seeds; error bars show across-seed SE.}
    \label{fig:scaling}
\end{figure}

\paragraph{Training set scaling.} Increasing the training set from 1\% to 100\% reduces $\E[T]$ by 19\%, from $14.67$ to $11.90$ expansions (\Figref{fig:scaling}a). With 75\% of the training data, search cost is already within 1\% of the full-data result. 

\paragraph{A Small Priority Model Retains the Signal.} Search with the deployed 0.6B Qwen3 model uses 7\% more expansions than with the 14B model ($11.87$ compared with $11.10$; \figref{fig:scaling}b). The deployed model has more than $20\times$ fewer parameters. We also evaluate models from five different families. Qwen3.5-2B uses 5\% fewer expansions than the deployed model ($11.31$) and outperforms larger models from other families. SmolLM2-360M uses 7\% more expansions ($12.70$).

\paragraph{Training-data difficulty and diversity.}
Deep repair trees provide effective supervision: training on the deep Qwen3-Coder-30B subset achieves a mean search cost of $8.291$ expansions across four evaluation proposers, compared with $8.290$ for the full corpus. Training on all four proposer corpora reduces mean cost by 3.5\%, from $8.29$ to $8.00$ expansions (\appref{app:transfer-analysis}).

\section{Related Work}
\label{sec:related}

\noindent\textbf{Adaptive search over stochastic program candidates.}
REx allocates refinement attempts with Thompson sampling and pass-rate-informed Beta posteriors \citep{tang2024code}.
AB-MCTS adaptively widens or deepens search using reward posteriors \citep{inoue2026wider}, while BG-MCTS conditions its search policy on the remaining budget \citep{miyamoto2026aligning}.
Their expansion rules use scalar evaluations and search-state statistics.
\fmethod{} learns its priorities from the parent state and proposed edit, using the repair's content to choose among unexpanded edges.

\noindent\textbf{Learned search priorities.}
Learned distance-to-go estimates guide DeepCubeA's heuristic search \citep{agostinelli2019solving}, while Q* uses action values to prioritize successors before generating them \citep{agostinelli2024q}.
For LLMs, process reward models score intermediate reasoning steps \citep{uesato2022solving,lightman2024let}, and outcome verifiers rank completed solutions \citep{cobbe2021training}.
Tree-of-Thoughts, RAP, and LATS use model-based evaluations to guide search \citep{yao2023tree,hao2023reasoning,zhou2024language}.
DREAM scores plans by predicted success \citep{cui2026reward}; \fmethod{} learns solve-distance distributions from depth-censored branches to prioritize proposed edits across the full frontier.

\noindent\textbf{Censored distance prediction.}
Discrete-time survival models learn from exact and censored event times \citep{tutz2016modeling}; our softmax-over-bins parameterization follows \citet{lee2018deephit}.
In depth-limited repair trees, a fix reveals a solve distance, while a branch with no fix within the available depth supplies a lower bound.
This provides supervision from both solved branches and branches with no observed fix.

\noindent\textbf{Code repair and execution feedback.}
CodeTree guides code-generation search with execution feedback and LLM critiques \citep{li2025codetree}, while SWE-Search combines MCTS with an LLM value agent for repository-level repair \citep{antoniades2025swe}.
Self-debugging and Reflexion use feedback to improve subsequent attempts \citep{chen2024teaching,shinn2023reflexion}, and \citet{olausson2024self} analyze when self-repair outperforms resampling.
\fmethod{} keeps the proposer fixed and learns which of its proposed edits to expand, so improvements in repair generation and expansion order can be combined.

\section{Limitations and Future Work}
\label{sec:limitations}

\noindent\textbf{Task and language scope.}
The experiments demonstrate generalization to held-out Python programming tasks and transferability across 11 unseen repair LLMs. The selection interface uses execution feedback and proposed-edit descriptions. Evaluating the learned policy in other programming languages and repository-scale repair would extend the demonstrated scope.

\noindent\textbf{Information sharing and longer horizons.}
Sharing information across branches or generating additional candidates in response to prior decisions falls outside the scope of this work. The value of deeper training trajectories (\appref{app:transfer-analysis}) motivates learning over longer repair horizons.

\section{Conclusion}
Efficient code-repair search requires deciding which proposed edit should receive the next expensive generation and test execution. \fmethod{} allows comparing the programs and proposed edits globally by learning their solve-distance probabilities. \fmethod{} requires 21\% fewer expansions and 12\% fewer output tokens than the strongest baseline. The same frozen 0.6B-parameter policy achieves the highest success rate averaged across budgets for each of eleven unseen proposers. Ablations show the value of both semantic inputs and of comparing candidates globally across different parents and depths.

These results identify descriptions of unexecuted actions as a useful control signal for test-time search. The proposer determines which repairs are available; semantic prioritization determines which should receive compute first. Separating these roles improves search without proposer-specific tuning and provides a practical way to allocate test-time computation in code-repair tasks.

\newpage

\subsection*{AI use statement}

We used generative AI tools to assist with manuscript editing and consistency checks. The authors take responsibility for the final text, claims, and artifacts.

\subsection*{Ethics statement}

The study uses programming problems and model-generated repair trajectories; it involves no human participants and no personal data. The planned corpus release will document its source benchmarks, applicable licenses, generation procedure, and filtering so downstream users can assess reuse constraints. As with other code-generation systems, the method may produce insecure or incorrect programs, so generated code should be tested and reviewed before use.

\subsection*{Reproducibility statement}

The main paper defines the search setting, loss, priority, metrics, split policy, checkpoint-selection protocol, and baselines. The appendix reports data construction, training hyperparameters, baseline sweeps, seed aggregation for within-method analyses, and full per-policy results. 

\bibliography{iclr2027_conference}
\bibliographystyle{iclr2027_conference}

\appendix
\section{Dataset and Splits}
\label{app:dataset}

\paragraph{Source and tree construction.}
We use the \texttt{hard} subset of KodCode-V1 \citep{xu2025kodcode}, approximately 8,700 Python programming problems.
Qwen3-Coder-30B \citep{yang2025qwen3} generates diagnoses and up to $k=3$ proposed edits per node and up to five initial solutions (roots) per problem, with maximum depth $D=4$.
Every candidate program is executed against the task's complete test suite; pass rate 1.0 defines a fix.
The resulting corpus contains 45,424 trees and 3,967,516 executed nodes; 42 trees in which some node received fewer than $k$ proposed edits are dropped, leaving 45,382 trees and 3,963,103 nodes.

\paragraph{Retention and negatives.}
We retain 32,687 trees (72.0\%, 2,458,256 nodes) with a reachable fix.
Of these, 11,683 originate from Codeforces \citep{codeforces}, 8,769 from TACO \citep{li2023taco}, 7,777 from Code Contests \citep{li2022competition}, and 4,458 from APPS \citep{hendrycks2021measuring}.
This filter makes the number of expansions to first fix finite and gives every retained root a first-fix ordering target; unsuccessful branches within a retained tree are kept, so training rows contain edges that did not produce a fix within their remaining depth.
Wholly unsolvable roots contain no first-fix ordering target, although they could be useful for calibration or semi-supervised representation learning.

\paragraph{Splits.}
A deterministic hash of the canonical task identifier assigns problems, not trees, to 70/15/15 train/validation/test splits.
KodCode provides each problem in two prompt styles: \emph{instruct} states the task in natural language, and \emph{complete} gives function signatures with test examples. Both styles of one problem share a canonical identifier and cannot cross splits.
The resulting solvable-root counts are 22,690/4,996/5,001 from 4,537/986/982 canonical tasks.
Training-fraction experiments use nested prefixes of the training tasks and never change the validation or test sets.

\subsection{Four-Proposer Corpus}
\label{app:transfer-corpus}

The full corpus contains 12,187,444 nodes from four proposers over the same 45,424 root programs.
For every \texttt{(question\_id, initial\_solution\_index)} pair, the starting program is identical across proposers, and all four corpora use the same task-level split: 31,711/6,862/6,851 roots for train/validation/test.
Each node contains a candidate program, its execution feedback, the proposer's diagnosis, and its proposed edits.
All trees use depth $D=4$ and branching factor $k=3$.

\begin{table}[h]
    \caption{\textbf{Four-proposer corpus.} Nodes, train-split nodes, and nodes per root count every record over all 45,424 roots, before the partial-tree filter. Solve rate is the fraction of the 44,794 roots with a usable starting state (the 630 error roots excluded) whose exhaustive tree contains a fix.}
    \label{tab:corpus-four}
    \centering
    \small
    \begin{tabular}{@{}lrrrr@{}}
    \toprule
    Proposer & Nodes & Train nodes & Nodes/root & Solve rate \\
    \midrule
    \modelicon{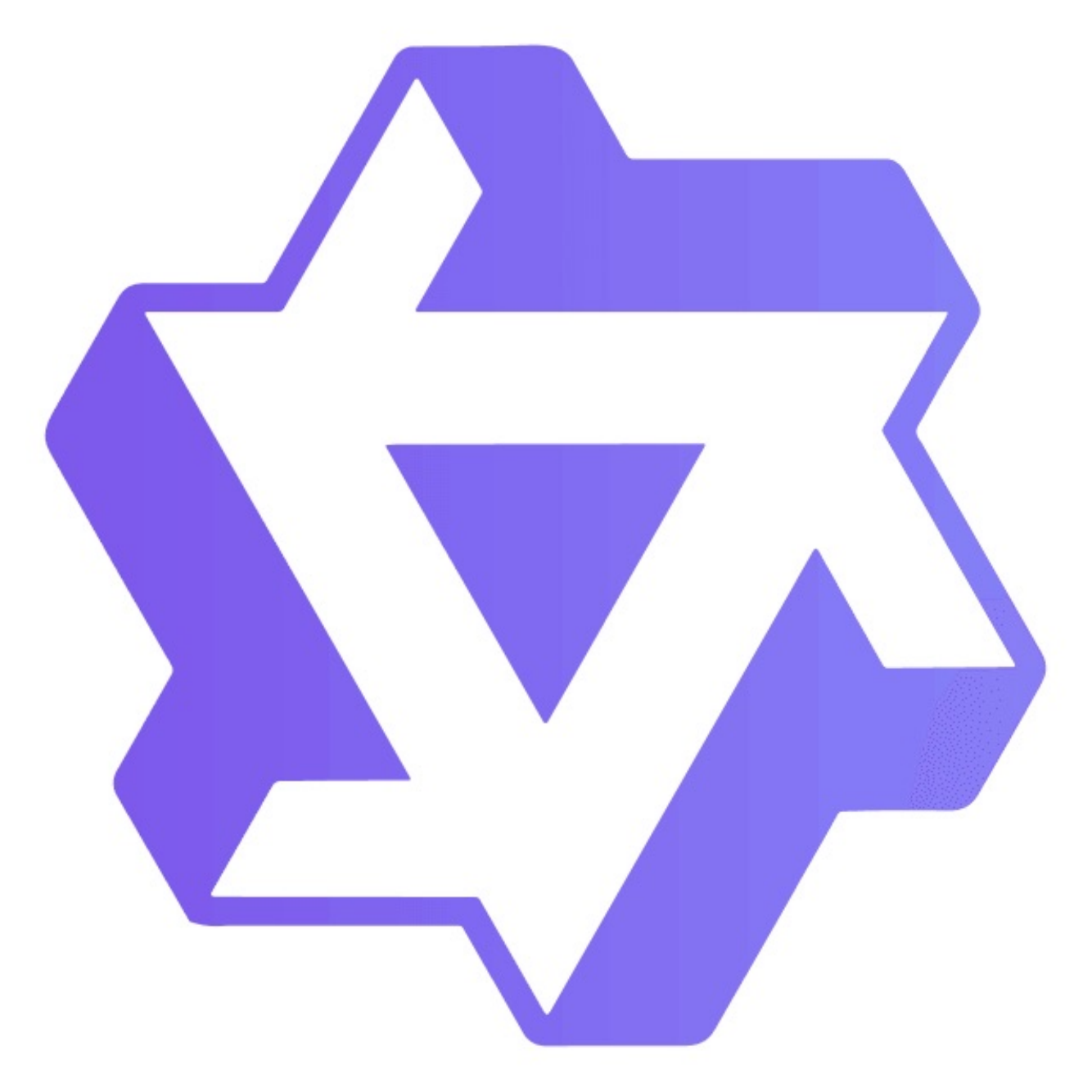}Qwen3-Coder-30B & 3,967,516 & 2,781,491 & 87.3 & 72.5\% \\
    \modelicon{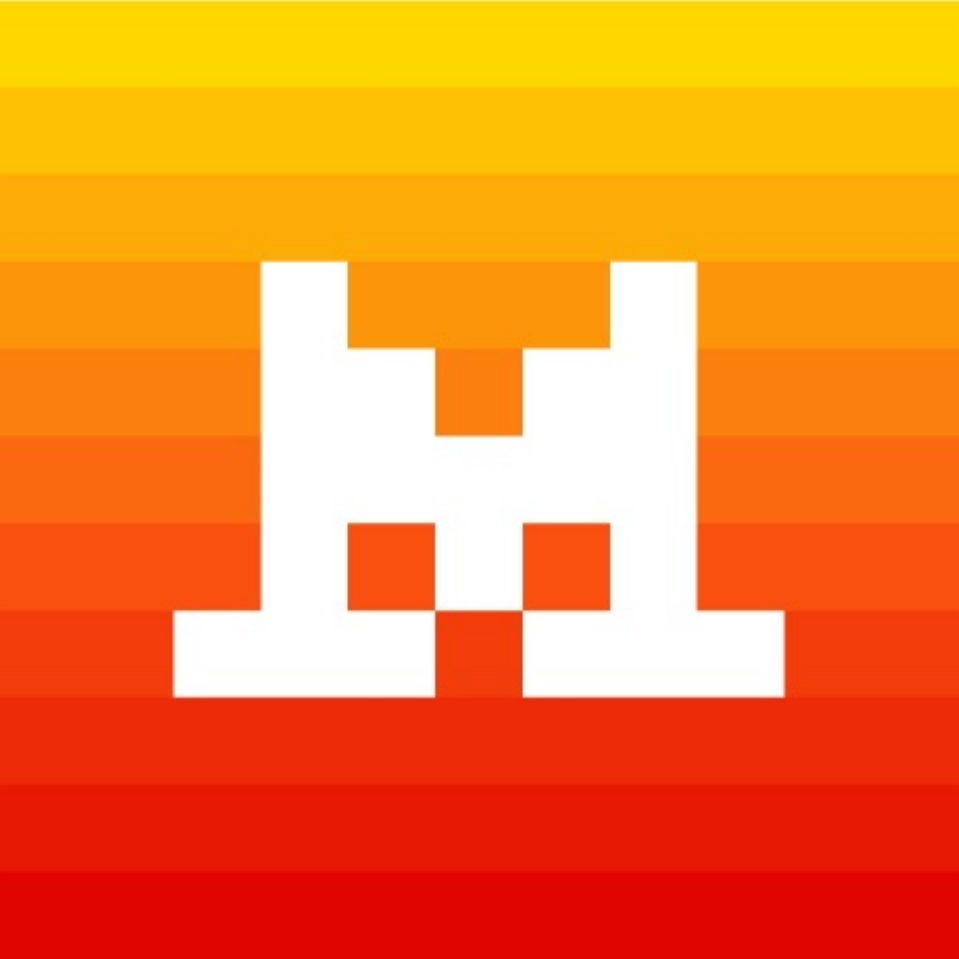}Devstral-Small-2 & 3,634,262 & 2,543,710 & 80.0 & 83.4\% \\
    \modelicon{qwen}Qwen3-Coder-Next & 2,329,481 & 1,633,236 & 51.3 & 91.6\% \\
    \modelicon{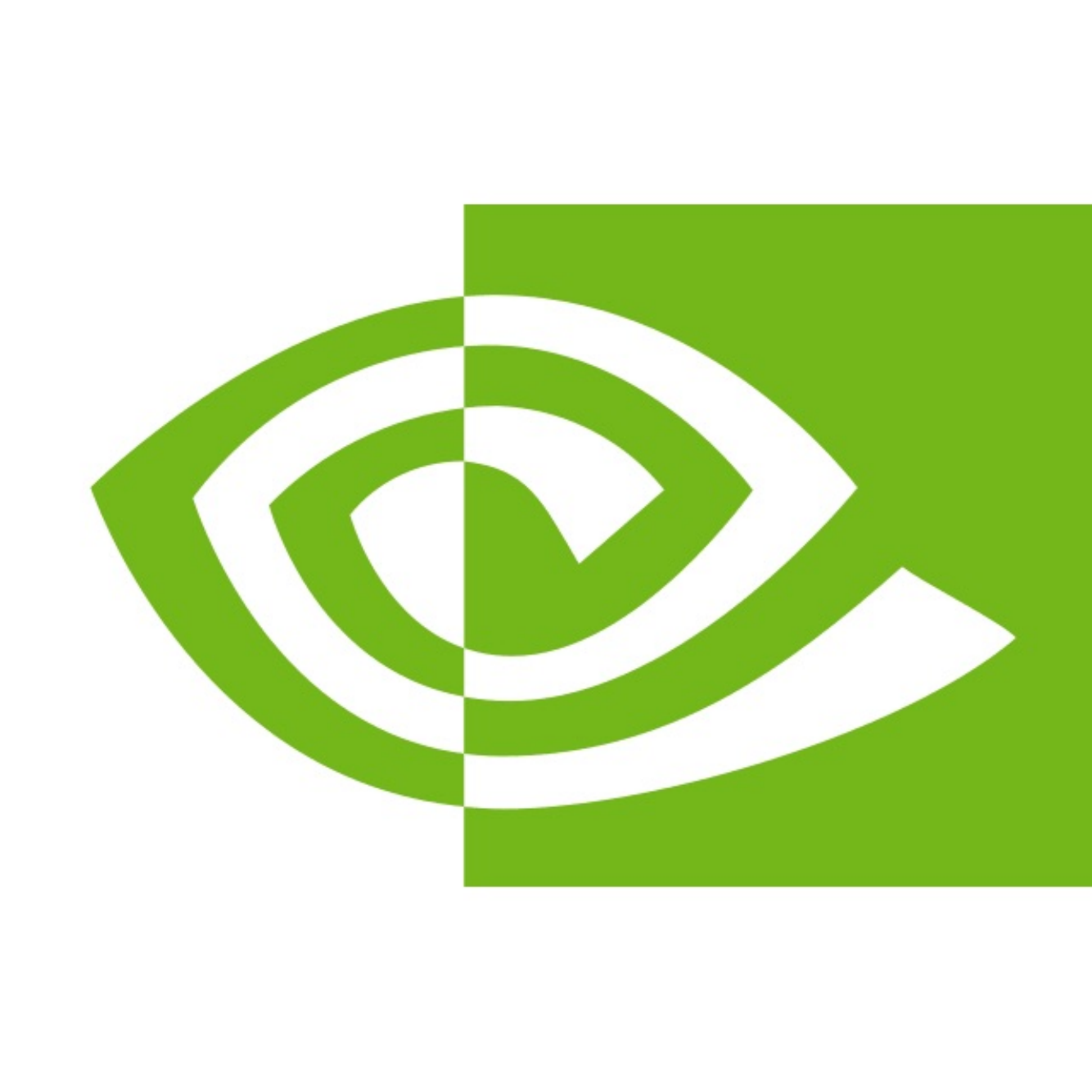}Nemotron-Cascade2 & 2,256,185 & 1,579,490 & 49.7 & 90.4\% \\
    \bottomrule
    \end{tabular}
\end{table}

\paragraph{Difficulty across proposers.}
For each root, we measure its solve distance, with roots that never solve placed last.
Pairwise Spearman correlations range from 0.52 to 0.68: programs that are difficult for one proposer tend to be difficult for the others, but proposer choice still changes a substantial part of the ordering.
The two Qwen3-Coder models have a correlation of 0.570, which is below the agreement observed between some models from different families at similar solve rates.

\begin{table}[h]
    \caption{\textbf{Root solve distance by proposer.} Percentages of the 44,794 roots with a usable starting state; roots already solved at the root are counted as solved but appear in no column. Stronger proposers solve more than half of their roots in one step, while Qwen3-Coder-30B produces more multi-step trajectories.}
    \label{tab:solve-depth}
    \centering
    \small
    \begin{tabular}{@{}lrrrrr@{}}
    \toprule
    Proposer & $t=1$ & $t=2$ & $t=3$ & $t=4$ & Never \\
    \midrule
    \modelicon{qwen}Qwen3-Coder-30B & 27.4\% & 18.9\% & 14.7\% & 11.5\% & 27.5\% \\
    \modelicon{mistral}Devstral-Small-2 & 34.7\% & 22.4\% & 15.9\% & 10.4\% & 16.6\% \\
    \modelicon{qwen}Qwen3-Coder-Next & 54.0\% & 21.7\% & 10.2\% & 5.7\% & 8.4\% \\
    \modelicon{nvidia}Nemotron-Cascade2 & 57.6\% & 18.7\% & 8.7\% & 5.5\% & 9.6\% \\
    \bottomrule
    \end{tabular}
\end{table}

\paragraph{Data integrity.}
There are 630 roots with an error diagnosis on the root record and no usable starting state; the solve statistics above exclude them.
They are flagged as the union across all four corpora, so filtering preserves the shared root set.
A further 2,443 trees contain an expanded node with one or two children because the proposer returned fewer than three proposed edits.
Loaders must pad these nodes rather than discard them; otherwise, 1.8\% of nodes are lost, concentrated in Qwen3-Coder-Next.

\section{Training-Proposer Transfer Analysis}
\label{app:transfer-analysis}

To study the effect of training data, we train separate priority models for the analyses in this section, evaluate on the validation split, and average three training seeds.
The evaluation population in each column contains the same solvable roots for every policy.

\begin{table}[h]
    \caption{\textbf{Transfer across training proposers.} Each row trains an identical priority model on trees from one proposer; columns report mean expansions to first fix. Lower is better, and bold marks matched training and evaluation proposers. Abbreviations: Q3-Coder, Qwen3-Coder-30B; Devstral-S2, Devstral-Small-2; Coder-Next, Qwen3-Coder-Next; Cascade2, Nemotron-Cascade2.}
    \label{tab:cross-proposer}
    \centering
    \small
    \setlength{\tabcolsep}{4pt}
    \begin{tabular}{@{}lrrrr@{}}
    \toprule
    Train & Q3-Coder & Devstral & Coder-Next & Cascade2 \\
    \midrule
    \modelicon{qwen}Q3-Coder-30B & \textbf{11.85} & 10.13 & 5.67 & 5.51 \\
    \modelicon{mistral}Devstral-S2 & 12.78 & \textbf{9.98} & 5.86 & 5.40 \\
    \modelicon{qwen}Q3-Coder-Next & 14.17 & 10.91 & \textbf{5.54} & 5.86 \\
    \modelicon{nvidia}Cascade2 & 13.99 & 11.51 & 6.24 & \textbf{5.25} \\
    \bottomrule
    \end{tabular}
\end{table}

\paragraph{Hard repairs distinguish training sources.}
Training-source differences emerge primarily on the hardest roots: across all twelve off-diagonal train--evaluation pairs, the mean penalty relative to matched training is $0.11$, $0.58$, $1.72$, and $3.33$ expansions as root solve distance increases from 1 to 4.

\begin{table}[h]
    \caption{\textbf{Transfer penalty on roots with solve distance 4.} Each cell is the change in expansions relative to the model trained on the evaluation corpus; negative values are improvements. The final column averages the other three evaluation corpora. Abbreviations as in \tabref{tab:cross-proposer}.}
    \label{tab:transfer-hard-tail}
    \centering
    \small
    \setlength{\tabcolsep}{3.5pt}
    \begin{tabular}{@{}lrrrrr@{}}
    \toprule
    Train & Q3-Coder & Devstral & Coder-Next & Cascade2 & Other mean \\
    \midrule
    \modelicon{qwen}Q3-Coder-30B & --- & $-0.15$ & $-0.11$ & $+0.22$ & \textbf{$-0.01$} \\
    \modelicon{mistral}Devstral-S2 & $+4.66$ & --- & $+4.13$ & $+2.36$ & $+3.72$ \\
    \modelicon{qwen}Q3-Coder-Next & $+6.83$ & $+1.15$ & --- & $+4.29$ & $+4.09$ \\
    \modelicon{nvidia}Cascade2 & $+5.57$ & $+5.20$ & $+5.83$ & --- & $+5.53$ \\
    \bottomrule
    \end{tabular}
\end{table}

Qwen3-Coder-30B is the least successful proposer but the strongest single-corpus training source: its model has a mean transfer penalty of $-0.01$ expansions on roots with solve distance 4 from the other corpora.
The corresponding penalties for the other training corpora range from $3.72$ to $5.53$ expansions.

\paragraph{Deep trajectories improve difficult repairs.}
We partition Qwen3-Coder-30B training trees into a shallow stratum of roots with solve distance 1 or 2 and a deep stratum of roots with solve distance 3 or 4 or no fix within the tree.
The deep stratum is capped at the shallow stratum's size, resulting in pools of 736,707 and 736,144 nodes, respectively.

\begin{table}[h]
    \caption{\textbf{Effect of deep rather than shallow supervision by root solve distance.} Entries are shallow minus deep $\E[T]$; positive values favor the deep training stratum.}
    \label{tab:depth-strata-by-distance}
    \centering
    \small
    \begin{tabular}{@{}lrrrr@{}}
    \toprule
    Evaluation proposer & $t=1$ & $t=2$ & $t=3$ & $t=4$ \\
    \midrule
    \modelicon{qwen}Qwen3-Coder-30B & $+0.02$ & $+0.30$ & $+0.64$ & $+1.84$ \\
    \modelicon{mistral}Devstral-Small-2 & $+0.03$ & $+0.31$ & $-0.15$ & $+1.99$ \\
    \modelicon{qwen}Qwen3-Coder-Next & $0.00$ & $-0.11$ & $+0.14$ & $+2.91$ \\
    \modelicon{nvidia}Nemotron-Cascade2 & $+0.01$ & $+0.04$ & $-0.74$ & $+0.38$ \\
    \bottomrule
    \end{tabular}
\end{table}

Deep supervision saves $0.38$--$2.91$ expansions over shallow supervision on roots with solve distance 4 across all four evaluation proposers. Costs are similar on immediately solved roots. The deep-stratum model also outperforms the model trained on Nemotron-Cascade2 on Devstral-Small-2 ($10.21$ versus $11.51$ expansions) and Qwen3-Coder-Next ($5.60$ versus $6.24$). Across the four evaluation sets, it achieves a mean cost of $8.291$ expansions, compared with $8.290$ for training on the full Qwen3-Coder-30B corpus.

\paragraph{Multi-proposer training.}
Training on all four corpora reduces mean validation search cost from $8.29$ to $8.00$ expansions. The resulting model achieves the lowest cost on three evaluation proposers; on the fourth, it is within $0.01$ expansions of the model trained on that proposer's corpus. The combined training pool contains 8,381,153 nodes after dropping trees with an incompletely expanded node, from the 8,537,927 training-split nodes in \tabref{tab:corpus-four}.

\begin{table}[h]
    \caption{\textbf{All-four mixture on validation.} Mean expansions to first fix; lower is better. Abbreviations as in \tabref{tab:cross-proposer}.}
    \label{tab:transfer-mixture}
    \centering
    \small
    \setlength{\tabcolsep}{4pt}
    \begin{tabular}{@{}lrrrrr@{}}
    \toprule
    Training corpus & \modelicon{qwen}Q3-Coder & \modelicon{mistral}Devstral & \modelicon{qwen}Coder-Next & \modelicon{nvidia}Cascade2 & Mean \\
    \midrule
    \modelicon{qwen}Qwen3-Coder-30B & 11.85 & 10.13 & 5.67 & 5.51 & 8.29 \\
    All four & \textbf{11.49} & \textbf{9.76} & 5.55 & \textbf{5.20} & \textbf{8.00} \\
    \bottomrule
    \end{tabular}
\end{table}

\section{Training and Architecture Details}
\label{app:training}

\paragraph{Backbone.}
The deployed priority model is \texttt{Qwen/Qwen3-0.6B-Base}, bf16, with LoRA adapters of rank 8, scale 16, and dropout 0.01 on the q/k/v/o projections; the base weights are frozen.
The backbone sweep applies the same recipe to Qwen3 1.7B/4B/8B/14B \citep{yang2025qwen3}, Qwen3.5 0.8B/2B \citep{qwen2026qwen35}, Llama-3.2 1B/3B \citep{meta2024llama32}, SmolLM2 360M/1.7B \citep{benallal2025smollm2}, and SmolLM3-3B \citep{huggingface2025smollm3}.

\paragraph{Input and head.}
One edge is serialized as parent execution feedback (at most 375 tokens), parent diagnosis (470), and edit text (175), separated by section markers.
The last-token hidden state is concatenated with four parent-level numeric features, standardized with training-corpus statistics that are stored in the checkpoint and never recomputed at deployment: test pass rate; parent depth divided by the maximum depth $D$; failed tests divided by passed plus failed tests (zero when no tests are recorded); and parent-code length in characters divided by the maximum code length in the training corpus. All four are constant across siblings. The edit text is therefore the only sibling-varying input to the language model, and the deployed model is not told the candidate's position in the proposer's list.
The head is Linear($\to256$)--GELU--dropout(0.1)--Linear($\to D{+}1$), with a zero-initialized output layer.

\paragraph{Optimization.}
We use AdamW with constant learning rates of $8\times10^{-5}$ for LoRA and $10^{-5}$ for the head, head-only weight decay of 0.01, gradient-norm clipping at 1.0, and an effective batch size of 64 across eight GPUs.
Each step samples one solvable parent and treats its child edges, including censored siblings, as a training row.
We evaluate the full best-first search on the validation set every 1,000 steps and stop after three evaluations without improvement.
Ablation configurations use three matched training seeds. The deployed configuration was trained with five seeds, and one checkpoint was selected based on validation performance before the headline test comparison.
Search uses deterministic \texttt{argmax} over frontier order; FIFO or uniform random tie-breaking gives identical validation costs, so results do not depend on tie handling.

\section{Baseline Adaptation and Tuning}
\label{app:baselines}
\label{app:tuning}

Under exhaustive-tree evaluation, requesting another child from REx, AB-MCTS, or BG-MCTS reveals the next pre-generated child below that parent.
The policy interface discloses a node's program, pass rate, execution feedback, and diagnosis only upon expansion; for each adaptive baseline, paired trees that differ only in the outcome of one unexpanded node confirm that the expansion sequence is identical up to and including the step that first expands that node.
Finite branching requires three mechanical adaptations: an exhausted REx arm leaves the pool; an exhausted AB-MCTS generation arm is removed; and BG-MCTS's completion bias, a bonus that grows with the depth of an unfinished generation chain, is identically zero because every revealed node is already a complete program.

We tune each adaptive policy on validation only.
REx sweeps $C_{\mathrm{REx}}\in\{10,20,40\}$ and uses $C_{\mathrm{REx}}=10$; BG-MCTS sweeps exploration $c\in\{1,\sqrt{2},2\}$ and widening $k_{\mathrm{BG}}\in\{2,4\}$ on a deterministic 512-root validation subsample with one seed. The gain comes entirely from $k_{\mathrm{BG}}=4$; $c=2$ scored within 0.1\% of $c=\sqrt{2}$, so we keep the reference $c=\sqrt{2}$ with $k_{\mathrm{BG}}=4$.
AB-MCTS retains its published conjugate priors: Beta$(0.5,0.5)$ and a Normal--Inverse-$\chi^2$ prior centered at 0.5.

\section{Complete Exhaustive-Tree Results}
\label{app:perseed}

\begin{table}[h]
    \caption{\textbf{Exhaustive-tree evaluation on the Qwen3-Coder-30B test split.} Mean expansions to first fix over 5,001 solvable roots from 982 canonical tasks; lower is better. The \fmethod{} checkpoint is selected on validation, while stochastic baselines average their inference-RNG runs. Uncertainty is a task-clustered bootstrap SE; $\Delta$ is paired against \fmethod{} on the same roots.}
    \label{tab:id-full}
    \centering
    \small
    \begin{tabular}{@{}lrrrr@{}}
    \toprule
    Policy & $\E[T]\downarrow$ & paired $\Delta$ & $z$ & Runs \\
    \midrule
    \textbf{\fmethod{}} & \textbf{$11.54$\se{0.41}} & --- & --- & 1 \\
    AB-MCTS-$\beta$ & $14.59$\se{0.48} & $+3.06$\se{0.24} & 12.9 & 3 \\
    AB-MCTS-$\mathcal N$ & $14.62$\se{0.48} & $+3.08$\se{0.24} & 12.8 & 3 \\
    REx & $14.92$\se{0.51} & $+3.39$\se{0.27} & 12.7 & 3 \\
    BG-MCTS & $15.67$\se{0.50} & $+4.13$\se{0.27} & 15.2 & 3 \\
    random & $16.54$\se{0.50} & $+5.00$\se{0.27} & 18.4 & 3 \\
    pass-rate & $17.01$\se{0.58} & $+5.47$\se{0.37} & 14.9 & 3 \\
    DFS & $17.01$\se{0.58} & $+5.47$\se{0.37} & 14.9 & 1 \\
    BFS & $17.14$\se{0.55} & $+5.61$\se{0.28} & 19.8 & 1 \\
    \bottomrule
    \end{tabular}
\end{table}

\paragraph{Output tokens to first fix.}
Let $G_i^\pi$ be the cumulative output tokens through the expansion that first fixes root $i$.
We average $G_i^\pi$ over the same 5,001 roots using the same selected-policy and RNG-averaged-baseline convention as the expansion comparison above.
Counts use the Qwen3-Coder-30B tokenizer and cover proposer output tokens; priority-model costs are reported in \appref{app:priority-cost}.

\begin{table}[h]
    \caption{\textbf{Output tokens to the first fix on the Qwen3-Coder-30B test split.} Mean output tokens, in thousands; lower is better. AB-MCTS reports its stronger variant on this metric (Gaussian posterior). The last column is the reduction obtained by \fmethod{} relative to each policy; uncertainty is a task-clustered bootstrap SE.}
    \label{tab:tokens-full}
    \centering
    \small
    \begin{tabular}{@{}lrr@{}}
    \toprule
    Policy & Tokens ($\mathrm{k}$)$\downarrow$ & \fmethod{} reduction \\
    \midrule
    \textbf{\fmethod{}} & \textbf{$14.96$\se{0.55}} & --- \\
    REx & $17.03$\se{0.60} & $12.1\%$ \\
    pass-rate & $17.18$\se{0.64} & $12.9\%$ \\
    DFS & $17.18$\se{0.64} & $12.9\%$ \\
    AB-MCTS & $18.12$\se{0.62} & $17.4\%$ \\
    random & $18.98$\se{0.61} & $21.2\%$ \\
    BG-MCTS & $19.95$\se{0.64} & $25.0\%$ \\
    BFS & $22.42$\se{0.74} & $33.3\%$ \\
    \bottomrule
    \end{tabular}
\end{table}

Pass-rate and DFS choose the same expansion sequence on 99.70\% of roots. Their remaining differences yield unrounded means of 17.007 and 17.008 expansions and 17,180 and 17,182 output tokens, respectively; the displayed table entries coincide after rounding.

\section{Mechanism Ablations}
\label{app:mechanism}

All results in this section use the 4,996-root validation split. Each neural variant averages three matched training seeds; uncertainty is the across-seed SE.

\begin{table}[ht]
    \caption{\textbf{Ablation summary.} Removing repair text or replacing the distributional priorities increases search cost. $\E[T]$ is the mean number of expansions to the first solution; lower is better. Each block compares against its own full-model reference in the first row; $\Delta$ is the increase in expansions. Values are the mean $\pm$ standard error across three matched training seeds. Priority-rule comparisons use the same trained distributional models as the objective block. Detailed input and objective controls follow below; the full priority-rule sweep is in \appref{app:reductions}.}
    \label{tab:ablations}
    \begin{center}
    \small
    \setlength{\tabcolsep}{4pt}
    \begin{tabular}{@{}llcc@{}}
    \toprule
    Axis & Variant & $\E[T]\downarrow$ & $\Delta$ \\
    \midrule
    Inputs & parent state $+$ proposed edit $+$ numerics & \textbf{$11.80$}\se{0.10} & --- \\
               & no parent-state text & $13.64$\se{0.11} & $+1.84$ \\
               & no edit text & $14.38$\se{0.09} & $+2.59$ \\
               & no text (matched model) & $17.14$\se{0.07} & $+5.35$ \\
    \midrule
    Objective & distributional solve distance & \textbf{$11.79$}\se{0.03} & --- \\
                & pairwise global-frontier ranking & $12.20$\se{0.08} & $+0.41$ \\
                & pairwise sibling ranking & $13.32$\se{0.08} & $+1.53$ \\
                & scalar distance-MSE & $13.59$\se{0.10} & $+1.80$ \\
                & scalar soft-BCE & $13.68$\se{0.15} & $+1.89$ \\
    \midrule
    Priority rule & normalized CDF area, \eqref{eq:priority} & \textbf{$11.79$}\se{0.03} & --- \\
             & one-step probability, $p_1$ & $13.58$\se{0.04} & $+1.78$ \\
             & rank separately within each depth & $13.96$\se{0.04} & $+2.16$ \\
    \bottomrule
    \end{tabular}
    \end{center}
\end{table}

\subsection{Inputs}

\begin{table}[h]
    \caption{\textbf{Input ablations.} The first six rows use the same priority model; the final two are numeric-MLP controls. Lower $\E[T]$ is better.}
    \label{tab:input-fields-full}
    \centering
    \small
    \begin{tabular}{@{}lrr@{}}
    \toprule
    Input & $\E[T]\downarrow$ & $\Delta$ from full \\
    \midrule
    parent state $+$ proposed edit $+$ numerics & \textbf{$11.80$\se{0.10}} & --- \\
    proposed edit $+$ numerics & $13.64$\se{0.11} & $+1.84$ \\
    proposed edit only & $13.65$\se{0.16} & $+1.85$ \\
    parent state $+$ numerics & $14.38$\se{0.09} & $+2.59$ \\
    parent state only & $14.39$\se{0.04} & $+2.60$ \\
    no text $+$ numerics & $17.14$\se{0.07} & $+5.35$ \\
    \midrule
    numeric MLP $+$ proposal-order feature & $14.45$\se{0.02} & $+2.66$ \\
    numeric MLP, no order feature & $16.79$\se{0.03} & $+5.00$ \\
    \bottomrule
    \end{tabular}
\end{table}

The four configurations with fixed numeric features isolate the contribution of parent-state text and proposed-edit text using the same backbone and training recipe. Parent-state text combines execution feedback and diagnosis, so attribution is to this combined field. Removing parent-state text increases cost by $1.84$ expansions, and removing the proposed edit increases it by $2.59$. Against the no-text policy, parent state alone saves $2.76$\se{0.09} expansions, edit text alone saves $3.50$\se{0.10}, and the full combination saves $5.35$\se{0.04}. The numeric-MLP control receives proposal order, a feature not available to the deployed policy; removing it increases cost from $14.45$ to $16.79$.

\subsection{Input Ablations under Proposer Shift}
\label{app:ood-inputs}

We evaluate the four representation-matched input configurations on exhaustive validation trees from three other proposers. Each row uses the same deterministic 2,000-root subsample for every variant.

\begin{table}[h]
    \caption{\textbf{Inputs under proposer shift.} Mean expansions to first fix; lower is better.}
    \label{tab:ood-inputs}
    \centering
    \small
    \setlength{\tabcolsep}{4pt}
    \begin{tabular}{@{}lrrrr@{}}
    \toprule
    Proposer & Full & No state & No edit & No text \\
    \midrule
    Devstral-Small-2 & \textbf{$10.11$\se{0.14}} & $12.28$\se{0.18} & $12.04$\se{0.18} & $14.23$\se{0.02} \\
    Nemotron-Cascade2 & \textbf{$5.85$\se{0.06}} & $6.78$\se{0.12} & $6.83$\se{0.04} & $8.16$\se{0.04} \\
    Qwen3-Coder-Next & \textbf{$5.27$\se{0.07}} & $6.06$\se{0.20} & $6.33$\se{0.08} & $7.67$\se{0.08} \\
    \bottomrule
    \end{tabular}
\end{table}

Both fields reduce cost for every proposer. Relative to no text, parent state alone saves 1.33 to 2.19 expansions and edit text alone saves 1.38 to 1.95. The full model is strongest on every proposer, showing that both semantic inputs remain useful beyond the model that generated the training trees.

\subsection{Objectives}

Distance-MSE regresses a sigmoid output onto the solve distance normalized to $[0,1]$, with censored edges assigned the maximum value $D{+}1$. Soft-BCE trains a sigmoid output on the probability that uniform random search below the edge reaches a fix, computed recursively as the mean over children (one for a fix, zero for a failed leaf). Pairwise sibling ranking uses a Bradley--Terry loss on a scalar score, preferring shorter solve distances and placing censored edges last. Global ranking uses the same loss with depth-normalized priorities as its targets, as defined below.

\begin{table}[h]
    \caption{\textbf{Objective and search cost.} Every variant uses the same backbone, inputs, optimizer, validation rule, and matched seeds. Sibling and global rankers share the scalar head and pairwise loss; their comparison sets and targets are described below.}
    \label{tab:objectives-full}
    \centering
    \small
    \begin{tabular}{@{}lrr@{}}
    \toprule
    Objective & $\E[T]\downarrow$ & $\Delta$ \\
    \midrule
    distributional solve distance & \textbf{$11.79$\se{0.03}} & --- \\
    pairwise global ranking, 3-edge sets & $12.20$\se{0.08} & $+0.41$ $(+3.4\%)$ \\
    pairwise global ranking, 8-edge sets & $12.17$\se{0.31} & $+0.38$ $(+3.2\%)$ \\
    pairwise sibling ranking & $13.32$\se{0.08} & $+1.53$ $(+13.0\%)$ \\
    scalar distance-MSE & $13.59$\se{0.10} & $+1.80$ $(+15.3\%)$ \\
    scalar soft-BCE & $13.68$\se{0.15} & $+1.89$ $(+16.0\%)$ \\
    \bottomrule
    \end{tabular}
\end{table}

\subsection{Global-Frontier Ranking}
\label{app:global-ranking}

The sibling and global rankers use the same scalar Bradley--Terry loss. The sibling variant compares the three children of one solvable parent; the global variant compares edits sampled from search frontiers on training trees. Frontiers come from one to eight expansions under a uniform mixture of breadth-first search, uniform random search, and choosing the earliest-listed available proposal, sampled independently of solve-distance labels. The primary variant samples up to three edges without replacement from each of eight frontiers per GPU and step. A secondary variant samples up to eight edges from each of three frontiers. Both allow up to 24 edges per step. In the primary variant, 81.7\% of sampled sets span multiple parents and 66.0\% span multiple depths.

The global training target is the \emph{oracle priority}: \eqref{eq:priority} evaluated with the observed solve distance. For an edge with remaining depth $b$, it is $u(e)=(b+1-t(e))/b$ when a fix is observed within $b$, and zero otherwise. At a shared depth, this preserves the sibling ranker's preference for shorter distances, with censored edges last. Across depths, it accounts for the available repair horizon. The loss averages over pairs with unequal targets within each set, then over sets containing at least one such pair.

Global comparisons reduce mean search cost from $13.32$ to $12.20$ expansions; the eight-edge variant reaches $12.17$. \fmethod{} achieves the lowest mean cost at $11.79$ expansions (\tabref{tab:objectives-full}). Global ranking recovers 73\% of the sibling ranker's gap to \fmethod{}, demonstrating the value of learning comparisons across the frontier. The remaining paired difference of $0.41$ expansions is not statistically resolved with three seeds.

Every model scores the same 4,050 validation choice sets, sampled with a separate random seed and containing up to eight edges with at least two distinct oracle priorities. \emph{Frontier precision at one} (P@1) measures how often the top-scored edit has maximal oracle priority. \emph{Frontier pairwise accuracy} measures correct ordering of pairs with unequal oracle priorities, giving tied predictions half credit. \emph{Cross-parent accuracy} applies this rule to edits proposed by different programs. Sibling P@1 is measured separately on groups with differing solve distances and counts selections with the smallest distance as correct.

\begin{table}[h]
    \caption{\textbf{Held-out frontier ordering.} \fmethod{} achieves the highest frontier P@1 and pairwise accuracy, alongside the lowest search cost. Ranking entries are percentages averaged over three seeds; P@1 and pairwise metrics are defined above. Lower $\E[T]$ is better.}
    \label{tab:frontier-ranking}
    \centering
    \small
    \setlength{\tabcolsep}{3pt}
    \begin{tabular}{@{}lrrrrr@{}}
    \toprule
    Objective & Sibling P@1 & Frontier P@1 & Frontier pairwise & Cross-parent & $\E[T]$ \\
    \midrule
    distributional solve distance & 61.1 & \textbf{43.1} & \textbf{66.5} & \textbf{67.4} & \textbf{11.79} \\
    pairwise global, 3-edge sets & 59.9 & 42.8 & 64.6 & 65.2 & 12.20 \\
    pairwise global, 8-edge sets & 59.6 & 42.2 & 64.6 & 65.3 & 12.17 \\
    pairwise sibling ranking & \textbf{61.4} & 38.7 & 58.1 & 57.5 & 13.32 \\
    \bottomrule
    \end{tabular}
\end{table}

\fmethod{} achieves the highest accuracy when comparing edits from different programs: 67.4\%, compared with 57.5\% for sibling ranking and 65.2--65.3\% for global ranking. Its stronger frontier comparisons accompany the lowest measured search cost. This connects the benefit of solve-distance prediction to the decisions the policy makes throughout repair search.

\subsection{Cheap Text Baseline}
\label{app:cheap-baseline}

We train a hashing TF--IDF plus logistic regression baseline on the same training split and score it with the same search evaluator. It recovers substantial value from text, but the priority model still costs 16\% less on identical inputs (2.18 expansions).

\begin{table}[h]
    \caption{\textbf{Cheap learned text baseline.} Lower $\E[T]$ is better.}
    \label{tab:tfidf}
    \centering
    \small
    \begin{tabular}{@{}lr@{}}
    \toprule
    Input and model & $\E[T]\downarrow$ \\
    \midrule
    TF--IDF: parent state $+$ edit $+$ numerics & 13.97 \\
    TF--IDF: proposed edit only & 14.10 \\
    TF--IDF: parent state only & 15.70 \\
    TF--IDF: numerics only & 16.52 \\
    \textit{0.6B priority model: same full input} & \textit{11.79} \\
    \bottomrule
    \end{tabular}
\end{table}

\section{Priority-Rule Ablations}
\label{app:reductions}

Let $A=\sum_{j\le b}p_j(b+1-j)=\sum_{m=1}^{b}F(m)$.
The deployed priority is $A/b$.
All rows below re-score the same three validation checkpoints; no model is retrained.

\begin{table}[ht]
    \caption{\textbf{Priority-rule sweep on validation.} Mean expansions to first fix, sorted by cost. Every row reuses the same predictions, roots, and deterministic tie-break; only the mapping from the predicted distribution to a priority changes.}
    \label{tab:reductions-full}
    \centering
    \small
    \begin{tabular}{@{}lrr@{}}
    \toprule
    Priority rule & $\E[T]\downarrow$ & $\Delta$ \\
    \midrule
    \textbf{CDF area, $A/b$} & \textbf{$11.79$\se{0.03}} & --- \\
    affine, $A/b+0.1F(b)$ & $11.81$\se{0.02} & $+0.02$ \\
    discounted, $\sum_{j\le b}\gamma^{\,j-1}p_j$, $\gamma=0.5$ & $11.87$\se{0.02} & $+0.08$ \\
    solve probability / capped distance, $F(b)/(b+1-A)$ & $11.88$\se{0.04} & $+0.08$ \\
    expected distance, $A/(b+1)$ & $12.00$\se{0.03} & $+0.20$ \\
    solve within remaining depth, $F(b)$ & $12.23$\se{0.05} & $+0.44$ \\
    CDF area, unnormalized $A$ & $12.71$\se{0.06} & $+0.92$ \\
    one-step probability, $p_1$ & $13.58$\se{0.04} & $+1.78$ \\
    per-depth ranks of CDF area & $13.96$\se{0.04} & $+2.16$ \\
    solve-distance median, $(b+1-q_{0.5})/(b+1)$ & $14.61$\se{0.10} & $+2.82$ \\
    \bottomrule
    \end{tabular}
\end{table}

Here $q_{0.5}$ is the first bin at which the CDF reaches $0.5$ within the remaining depth, or $b+1$ if it never does. The denominator $b+1-A=\E[\min(t,b+1)]$ is the expected solve distance capped at $b+1$. At fixed depth, $A$, $A/b$, and $A/(b+1)$ rank edges identically; their difference is how edges at different depths interleave in one queue.
No tested rule improves on $A/b$. Several alternative timing-aware rules fall within $0.10$ expansions. Removing the depth normalizer $1/b$ increases search cost by $0.92$ expansions. Replacing score magnitudes with separate within-depth ranks increases cost by $2.16$ expansions. These results show that score magnitudes help compare edits across depths, beyond ordering edits at the same depth.

\section{Scaling Tables}
\label{app:scaling}

\begin{center}
    \small
    \textbf{Training-data scaling on validation.}\\[3pt]
    \begin{tabular}{@{}lrrrrrrrr@{}}
    \toprule
    Training fraction & 1\% & 2\% & 5\% & 12.5\% & 25\% & 50\% & 75\% & 100\% \\
    \midrule
    $\E[T]$ & 14.67 & 14.26 & 13.48 & 13.15 & 12.50 & 12.31 & 12.02 & 11.90 \\
    \bottomrule
    \end{tabular}
\end{center}

\begin{center}
    \small
    \textbf{Backbone scaling on validation.} Every model uses the same distributional objective and priority.\\[3pt]
    \setlength{\tabcolsep}{2pt}
    \begin{tabular}{@{}lrrrrrrrrrrrr@{}}
    \toprule
    Backbone & Q3-.6 & Q3-1.7 & Q3-4 & Q3-8 & Q3-14 & Q3.5-.8 & Q3.5-2 & L3.2-1 & L3.2-3 & S2-.36 & S2-1.7 & S3-3 \\
    \midrule
    $\E[T]$ & 11.87 & 11.72 & 11.35 & 11.19 & 11.10 & 11.57 & 11.31 & 11.95 & 11.50 & 12.70 & 12.16 & 11.40 \\
    \bottomrule
    \end{tabular}
\end{center}

\section{Cross-Proposer Evaluation Protocol and Results}
\label{app:online}

\paragraph{Evaluation population.}
We train and select the priority model using only Qwen3-Coder-30B data, then deploy it unchanged with eleven repair LLMs: Devstral-Small-2, Gemma-4-31B, Nemotron-3-Nano, Qwen3-Next-80B, Nemotron-Cascade2, Qwen3-Coder-Next, Nemotron-3.5-Lightning, Qwen3.6-27B, GLM-5.2, GPT-OSS-20B, and Nemotron-3-Super \citep{mistral2025devstral2,gemmateam2026gemma4,nvidia2025nemotronnano,qwen2025qwen3next,yang2026cascade,qwen2026qwen3codernext,nvidia2026lightning,qwen2026qwen36,zeng2026glm5,openai2025gptoss,nvidia2026nemotronsuper}.
The three corpus-based evaluations (Devstral-Small-2, Nemotron-Cascade2, and Qwen3-Coder-Next) contribute 6,470 roots each; the eight online campaigns contribute 6,851 roots each. All use initial-solution indices 0--4 and an expansion budget of $B=36$, without screening for a reachable fix.
We average across all eleven corpora, giving each proposer equal weight.

\paragraph{Paired comparisons.}
For each proposer, we compare \fmethod{} with the adaptive baseline having the highest $\ausc(36)$ and hold that comparator fixed during resampling. We bootstrap canonical tasks with replacement (10,000 replicates), keeping each task's initial solutions together and evaluating both policies on the same sampled tasks. Proposers evaluated on the same task set share resampling draws. We report pointwise 95\% intervals for the fixed checkpoint and proposer set, without adjustment for comparator selection or multiple comparisons.

\begin{table}[ht]
    \caption{\textbf{Transfer with one frozen priority model.} $\ausc(36)$ is reported as a percentage. Each row names the strongest baseline for that proposer; $\Delta$ is the gain in percentage points, with 95\% task-clustered paired-bootstrap intervals in grey. All rows use the same Qwen3-Coder-30B-trained priority model.}
    \label{tab:online}
    \centering
    \small
    \setlength{\tabcolsep}{5pt}
    \begin{tabular}{@{}lccc@{}}
    \toprule
    Proposer &  baseline & \fmethod{} & $\Delta$ \\
    \midrule
    \modelicon{mistral}Devstral-Small-2       & 65.54 (REx)     & \textcolor{nvidia_green}{67.66} & $+2.13$\ci{+1.56}{+2.69} \\
    \modelicon{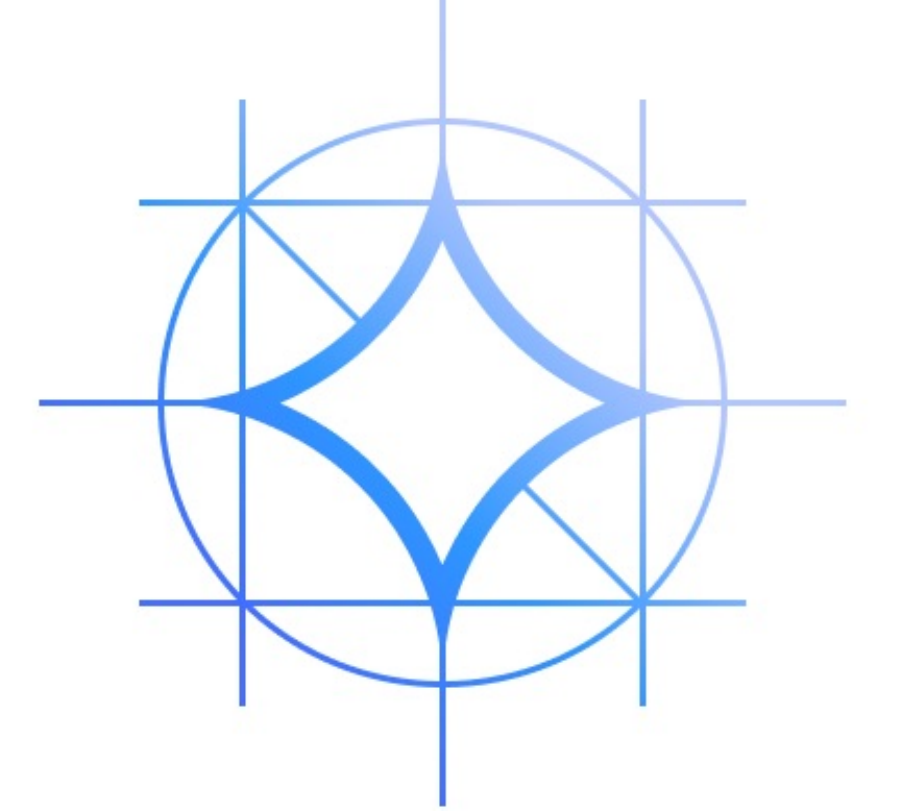}Gemma-4-31B              & 74.92 (AB-MCTS) & \textcolor{nvidia_green}{76.34} & $+1.41$\ci{+0.98}{+1.86} \\
    \modelicon{nvidia}Nemotron-3-Nano          & 77.48 (AB-MCTS) & \textcolor{nvidia_green}{78.12} & $+0.64$\ci{+0.30}{+0.99} \\
    \modelicon{qwen}Qwen3-Next-80B             & 79.32 (REx)     & \textcolor{nvidia_green}{80.64} & $+1.32$\ci{+0.90}{+1.74} \\
    \modelicon{nvidia}Nemotron-Cascade2        & 81.37 (REx)     & \textcolor{nvidia_green}{82.29} & $+0.91$\ci{+0.51}{+1.31} \\
    \modelicon{qwen}Qwen3-Coder-Next           & 82.10 (REx)     & \textcolor{nvidia_green}{83.50} & $+1.40$\ci{+0.99}{+1.82} \\
    \modelicon{nvidia}Nemotron-3.5-Lightning   & 84.44 (AB-MCTS) & \textcolor{nvidia_green}{85.40} & $+0.96$\ci{+0.61}{+1.32} \\
    \modelicon{qwen}Qwen3.6-27B                & 86.25 (AB-MCTS) & \textcolor{nvidia_green}{86.78} & $+0.53$\ci{+0.20}{+0.84} \\
    \modelicon{nvidia}Nemotron-3-Super         & 87.60 (AB-MCTS) & \textcolor{nvidia_green}{88.01} & $+0.41$\ci{+0.07}{+0.75} \\
    \modelicon{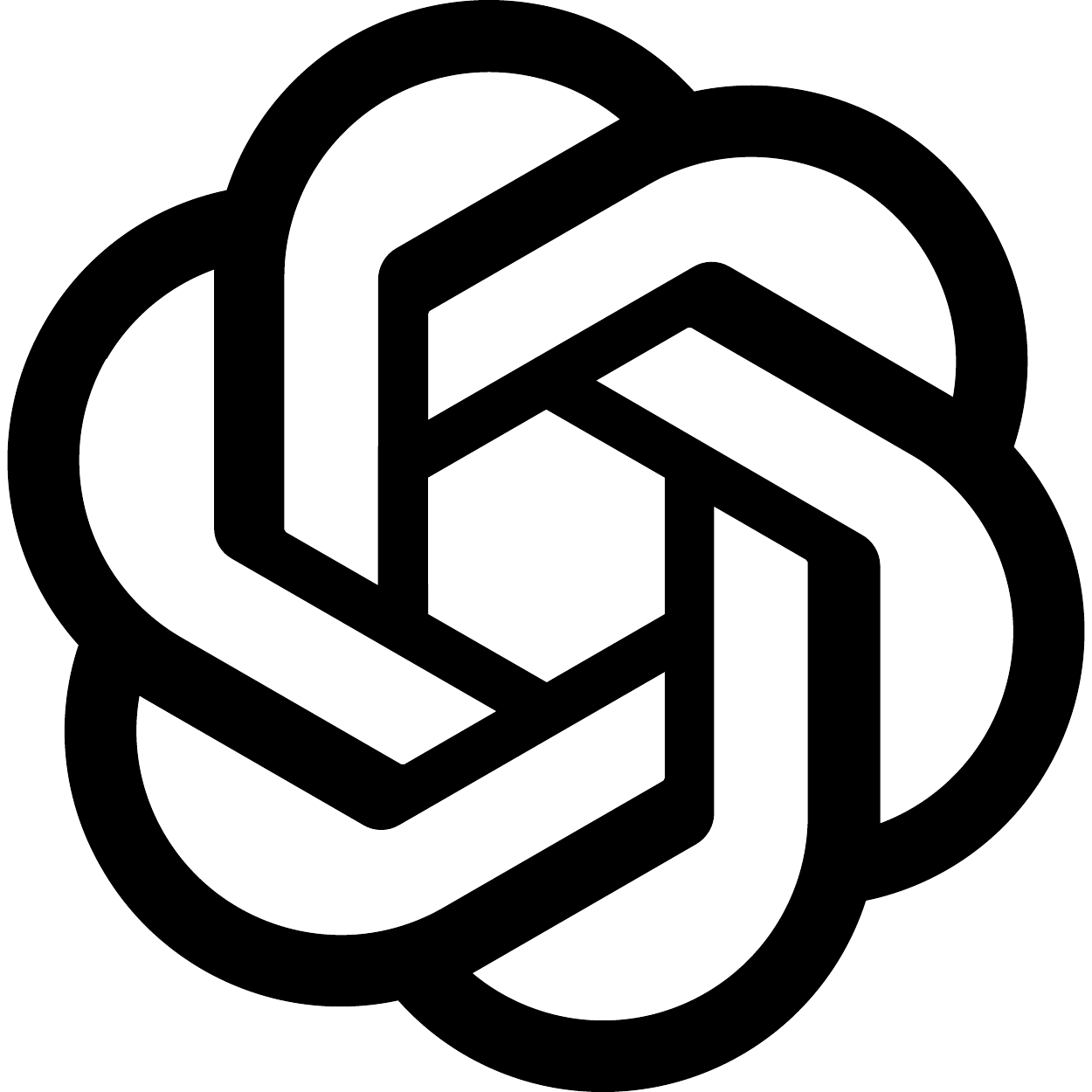}GPT-OSS-20B                         & 88.27 (REx)     & \textcolor{nvidia_green}{88.49} & $+0.22$\ci{-0.19}{+0.62} \\
    \modelicon{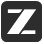}GLM-5.2                             & 90.32 (AB-MCTS) & \textcolor{nvidia_green}{90.33} & $+0.01$\ci{-0.25}{+0.27} \\
    \midrule
    \textbf{Mean (11 proposers)} & \textbf{81.60 (per-proposer best)} & \textcolor{nvidia_green}{\textbf{82.50}} & $\mathbf{+0.90}$\ci{+0.77}{+1.04} \\
    \bottomrule
    \end{tabular}
\end{table}

\paragraph{Budget sweeps.}
Reading a native-$B=36$ trajectory at a smaller budget is exact for every method except BG-MCTS, whose annealing schedule consumes the declared total budget.
Accordingly, sub-36 BG-MCTS points are approximate and are faded in the figures.
\fmethod{} depends on remaining depth, not the expansion budget $B$, so its truncated curves are exact.

The two aggregate views in \tabref{tab:online-mean} are algebraically linked: $\ausc(B)=1+S(B)/B-C(B)/B$ exactly.
We select the strongest adaptive baseline separately for each proposer and metric. Relative to these baselines, \fmethod{} improves AUSC by 1.1\% and success by 0.5\%, averaged across all eleven proposers. It also uses 4.3\% fewer capped expansions at $B=36$. The absolute changes are $+0.0090$, $+0.0040$, and $-0.3193$, respectively.

\begin{table}[!ht]
    \caption{\textbf{Mean performance across all eleven proposers.} Each proposer receives equal weight. Entries are success percentage / mean capped expansions; failures consume the full budget. The final column reports $\ausc(36)$ as a percentage.}
    \label{tab:online-mean}
    \centering
    \small
    \setlength{\tabcolsep}{3.5pt}
    \begin{tabular}{@{}lrrrrrrr@{}}
    \toprule
    Policy & $B=1$ & $B=4$ & $B=8$ & $B=16$ & $B=24$ & $B=36$ & $\ausc(36)$ \\
    \midrule
    \textbf{\fmethod{}} & \textbf{49.0/1.00} & \textbf{72.1/2.21} & \textbf{79.7/3.19} & \textbf{84.8/4.60} & \textbf{87.0/5.73} & \textbf{88.9/7.18} & \textbf{82.50} \\
    REx & 49.0/1.00 & 70.3/2.25 & 78.4/3.29 & 83.8/4.79 & 86.1/5.99 & 88.1/7.54 & 81.48 \\
    AB-MCTS-$\beta$ & 49.0/1.00 & 69.9/2.26 & 78.1/3.32 & 83.8/4.83 & 86.3/6.03 & 88.4/7.54 & 81.49 \\
    AB-MCTS-$\mathcal N$ & 49.0/1.00 & 69.8/2.26 & 78.0/3.33 & 83.8/4.84 & 86.4/6.03 & 88.4/7.54 & 81.50 \\
    BG-MCTS & 49.0/1.00 & 69.0/2.37 & 75.9/3.49 & 82.8/5.17 & 85.5/6.44 & 87.6/8.05 & 80.06 \\
    \bottomrule
    \end{tabular}
\end{table}

\paragraph{Early-budget gains.}
By the second expansion, \fmethod{} has higher success than REx and both AB-MCTS variants on all eleven proposers. The gains emerge early in the search.

\begin{figure}[!htp]
    \centering
    \includegraphics[width=0.95\textwidth,
        alt={Grid of twelve plots of success rate against mean expansions, one for each repair LLM: Qwen3-Coder-30B and eleven unseen LLMs. In each plot, the CodeTriage curve lies at or above the curves for REx, AB-MCTS, BG-MCTS, DFS, and BFS.}]{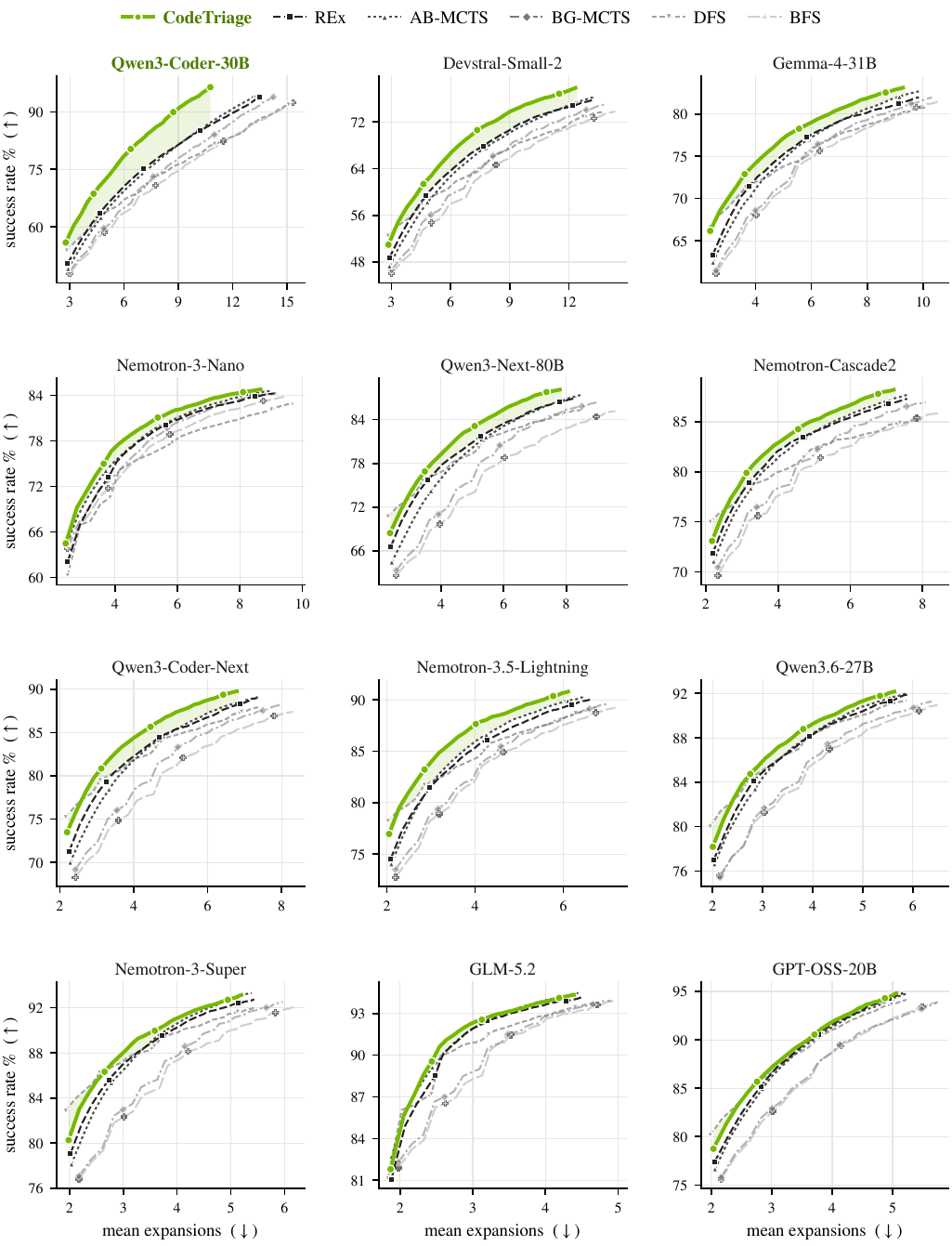}
    \caption{\textbf{Success--cost curves by proposer.}
    Success rate plotted against mean capped expansions as the expansion budget varies; failures consume the full budget.
    The first panel is the exhaustive-tree evaluation with Qwen3-Coder-30B, the training proposer.
    The next eleven panels show deployment with unseen proposers across a wide range of proposer capabilities. \fmethod{} achieves the strongest $\ausc(36)$ in every panel.
    Sub-$36$ BG-MCTS points are approximate because its schedule depends on the declared total budget.}
    \label{fig:pareto-all}
\end{figure}

\clearpage

\section{Calibration}
\label{app:diagnostics}

\begin{figure}[ht]
    \centering
    \includegraphics[width=0.88\textwidth,
        alt={Two reliability diagrams of observed rate against predicted probability. The CodeTriage model trained with the censored loss lies closer to the diagonal than a model trained with an uncensored loss. Its expected calibration error is 0.027 against 0.069 for solve timing, and 0.068 against 0.172 for solving within the budget.}]{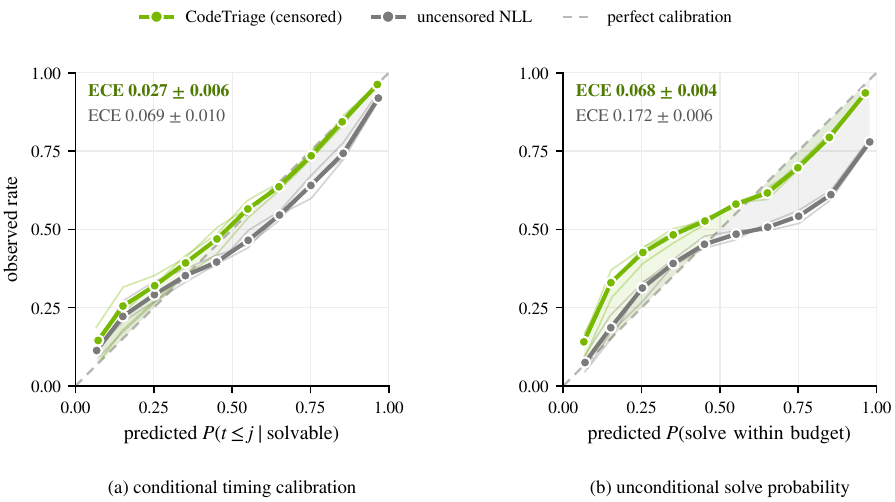}
    \caption{\textbf{What censoring calibrates.}
    Reliability of \textbf{(a)} the probability of solving within $j$ expansions, conditional on solving within the remaining depth, and \textbf{(b)} the unconditional probability of solving within that depth.
    Bold curves average three matched training seeds; faint curves show individual seeds, and the dashed diagonal is perfect calibration.
    Reported ECE is the mean of per-seed ECEs, with across-seed SE.}
    \label{fig:calibration}
\end{figure}

We compare the full model with an otherwise identical model trained only on edges with observed solve distances. The control drops censored edges from the loss instead of using the survival term in \eqref{eq:loss}. Across 44,964 validation predictions from three matched seeds, including censored edges reduces conditional timing ECE from $0.069$\se{0.010} to $0.027$\se{0.006} ($2.57\times$). It reduces unconditional solve-within-depth ECE from $0.172$\se{0.006} to $0.068$\se{0.004} ($2.54\times$). All three seeds improve on both measures (\figref{fig:calibration}). The paired improvements are $0.042$\se{0.012} and $0.105$\se{0.009}, respectively.

\clearpage
\section{Priority-Model Inference Cost}
\label{app:priority-cost}

Sharing the parent state across sibling edits increases priority-model throughput by $2.96\times$, from 123 to 363 edges/s on one H100.
At this throughput, scoring three proposed edits costs an amortized \textbf{8.3\,ms}, with \textbf{3.7\,GB} peak device memory.
The computation estimate below places scoring for a complete search at about half the proposer cost of one expansion, compared with 3.06 expansions saved against the strongest adaptive baseline.

\paragraph{Scoring workload.}
The 0.60B-parameter priority model (Qwen3-0.6B-Base with rank-8 LoRA and a 265K-parameter head) reads the parent state followed by one proposed edit and is pooled at a final \texttt{EOS} token.
Each edge is scored once and its priority cached: the parent state, proposed edit, and remaining tree depth are fixed for that edge.
A replay on a fixed subsample of 400 validation roots scores 10,464 edges over 4,872 expansions, or 2.15 edges per expansion on average.
Terminal programs and depth-$D$ leaves require no further scoring.
The per-edge input averages 789 tokens: 693.4 for the shared parent state, 94.6 for the edit, and one \texttt{EOS} token.

\paragraph{Sharing the parent state.}
The causal encoder allows the $k=3$ sibling edits to reuse the same state prefix.
We pack them into one row, $\mathrm{state}\,\|\,a_1\,\texttt{EOS}\,\|\,\cdots\,\|\,a_k\,\texttt{EOS}$, with a block-causal attention mask: each edit attends to the prefix and to itself, and rotary positions restart at the prefix length for each edit.
This preserves the per-edge computation in exact arithmetic and reduces the mean input for a full sibling group from 2,367 to approximately 980 tokens, a factor of 2.41.
In fp32, priorities differ by at most $3.7\times10^{-6}$ and the selected sibling agrees on all 1,500 checked parents.
In bf16, rounding can change near-tied priorities; both layouts achieve the same mean $\E[T]=12.18$ on the 400-root replay, with identical expansion traces on 75\% of roots.

\begin{table}[H]
    \caption{\textbf{Priority-model throughput on one H100 80GB HBM3 in bf16.} Per-edge scoring uses a separate row for each edit; prefix sharing uses one row per parent. Batch size counts edges, with parent counts shown for prefix sharing. The time for three edges is amortized from batched throughput; measurements cover priority scoring only.}
    \label{tab:priority-cost}
    \centering
    \small
    \begin{tabular}{@{}llrrr@{}}
    \toprule
    Layout & Batch (edges) & Edges/s & ms/3 edges & Peak memory \\
    \midrule
    Per edge & 32 & 122.1 & 24.6 & 2.2 GB \\
    Per edge & 128 & 122.6 & 24.5 & 5.2 GB \\
    Per edge & 512 & 118.0 & 25.4 & 17.0 GB \\
    Prefix-shared & 16$\times$3 & 344.9 & 8.7 & 1.9 GB \\
    Prefix-shared & 32$\times$3 & 358.2 & 8.4 & 2.5 GB \\
    Prefix-shared & 64$\times$3 & \textbf{363.4} & \textbf{8.3} & 3.7 GB \\
    Prefix-shared & 128$\times$3 & 361.3 & 8.3 & 6.0 GB \\
    \bottomrule
    \end{tabular}
\end{table}

\paragraph{Estimated compute relative to saved expansions.}
We estimate model matrix-multiplication cost as $2N_{\mathrm{active}}$ FLOPs per token.
Re-rendering the three proposer prompts issued per expansion gives 3,263 prefill tokens and 1,149 output tokens under \fmethod{}'s observed depth mix.
Charging the scorer's full 0.599B parameters and a full three-edge group of 980 tokens per expansion gives a scoring cost of approximately 3.7--4.4\% of proposer FLOPs for 3--3.6B-active MoE models and 0.43--0.49\% for dense 27--31B models.
Using the upper end of this range and the test-split mean of 11.54 expansions, scoring a complete search costs the equivalent of about \textbf{0.51 proposer expansions}.
\fmethod{} saves \textbf{3.06 expansions} against AB-MCTS-$\beta$, the strongest adaptive baseline (\tabref{tab:id-full}).
At the same mean proposer cost per expansion, the avoided generation therefore amounts to roughly \textbf{six times the scoring cost}, before counting the saved test executions.

\clearpage
\raggedbottom
\section{Qualitative Repair Example}
\label{app:qualitative}

This held-out Qwen3-Coder-30B search illustrates how \fmethod{} finds a reference-consistent repair through a branch with lower initial pass rates than its siblings.

\newtcblisting{codebox}[1][]{listing only,colback=black!2,colframe=black!30,boxrule=.4pt,arc=2pt,left=6pt,right=6pt,top=3pt,bottom=3pt,listing options={language=Python,basicstyle=\ttfamily\scriptsize,columns=fullflexible,keepspaces=true,showstringspaces=false,breaklines=true},#1}

\subsection{Recovering the Time and Capacity Constraints}
\label{app:case-cc39361}

\fmethod{} finds a reference-consistent repair in \textbf{14 expansions}, compared with \textbf{40.7} for the strongest adaptive baseline on this tree (AB-MCTS-$\mathcal N$, averaged over three inference seeds).
The task, \texttt{Code\_Contests\_39361\_C}, is supplied as the signature and docstring below together with seven tests.
Those tests require a shared time budget $T$ and at most $m$ completed tasks.

\begin{codebox}
def max_tasks_completed(n: int, T: int, task_durations: List[int], m: int) -> int:
    """
    Determines the maximum number of tasks that can be completed within a given total time
    by scheduling the tasks optimally among the available adventurers.
    """
\end{codebox}

Initial solution 1 (root 1007204) binary-searches the number of tasks and checks feasibility with a separate time budget per adventurer; it passes $3/7$ tests.

\begin{codebox}
from typing import List

def max_tasks_completed(n: int, T: int, task_durations: List[int], m: int) -> int:
    # Binary search on the answer
    # We want to find the maximum number of tasks that can be completed
    left, right = 0, n
    result = 0

    while left <= right:
        mid = (left + right) // 2
        if is_feasible(task_durations, m, T, mid):
            result = mid
            left = mid + 1
        else:
            right = mid - 1

    return result

def is_feasible(task_durations, m, T, k):
    # Check if we can complete exactly k tasks within time T with m adventurers
    if k == 0:
        return True

    # Sort durations in descending order to try to fit the longest tasks first
    durations = sorted(task_durations[:k], reverse=True)

    # Use greedy assignment to check if it's possible
    adventurers = [0] * m

    for duration in durations:
        # Find the adventurer with minimum current workload
        min_idx = adventurers.index(min(adventurers))
        adventurers[min_idx] += duration

        # If any adventurer exceeds time T, this assignment doesn't work
        if adventurers[min_idx] > T:
            return False

    return True
\end{codebox}

\begin{figure}[H]
    \centering
    \includegraphics[width=\textwidth,
        alt={Search tree for one repair task. CodeTriage expands 14 edits across several branches. It then follows a four-step path from the root program, which passes 3 of 7 tests, through direct greedy selection, sequential scheduling, greedy knapsack, and capping the count at m, to a program that passes all 7 tests. The best adaptive baseline needs 40.7 expansions on average.}]{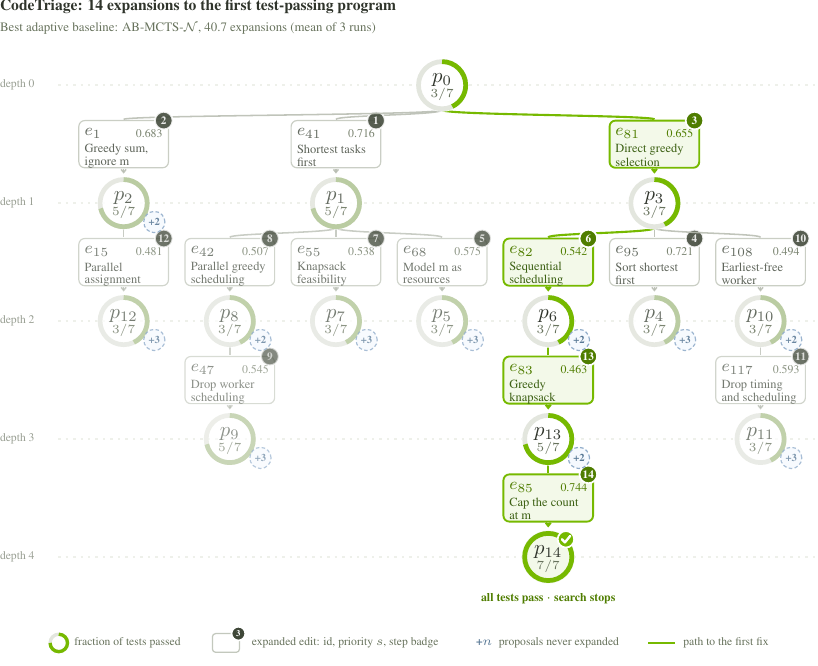}
    \caption{\textbf{Finding a repair beyond immediate test gains.}
    Programs $p_i$ are numbered in expansion order; rings show the fraction of tests passed.
    Each expanded edit $e_j$ ($j$ is the child's node id in the exhaustive tree) is a card with its recorded priority $s$ and the step at which it was expanded; $+n$ counts proposals never expanded.
    Green marks the path to the first passing program, identified retrospectively.
    The header compares \fmethod{} with the best adaptive baseline's mean over three inference seeds.}
    \label{fig:case-cc39361}
\end{figure}

The two depth-1 programs passing $5/7$ tests have no passing descendant within $D=4$.
The remaining sibling passes only $3/7$, but its branch contains both passing programs in the tree.
\fmethod{} returns to this branch while comparing edits across the frontier and completes the path $p_3\to p_6\to p_{13}\to p_{14}$.
The final program sorts task durations, enforces a shared time budget, and caps the number completed at $m$, matching the reference's selection rule.
This example demonstrates the value of searching beyond immediate pass-rate improvements.

\end{document}